\UseRawInputEncoding
\documentclass[uplatex,english, a4paper]{article}    

\usepackage{authblk}
\title{Estimation error analysis of deep learning on the regression problem on the variable exponent Besov space}
\date{}

\author[1, *]{Kazuma Tsuji}
\author[1, 2, **]{\ Taiji Suzuki}

\affil[1]{\it{Graduate School of Information Science and Technology, The University of Tokyo, Japan}}
\affil[2]{\it{Center for Advanced Intelligence Project, RIKEN, Japan}}

{
    \makeatletter
    \renewcommand\AB@affilsepx{: \protect\Affilfont}
    \makeatother

    \affil[ ]{Email}

    \makeatletter
    \renewcommand\AB@affilsepx{, \protect\Affilfont}
    \makeatother

    \affil[*]{kazuma\_tsuji@mist.i.u-tokyo.ac.jp}
    \affil[**]{taiji@mist.i.u-tokyo.ac.jp}
}

\usepackage[english]{babel}
\usepackage{a4wide}

\usepackage{amsthm}
\usepackage{mathtools}
\usepackage{amsmath}
\usepackage[pdftex]{graphicx}
\usepackage{listings}
\usepackage{here}
\usepackage{bm}
\usepackage{bbm}
\usepackage{comment}
\usepackage{mathrsfs}
\usepackage{tabularx}
 \usepackage{multicol}
\usepackage[subrefformat=parens]{subcaption}

\usepackage{natbib}
\usepackage{apalike}

\usepackage{tikz}
\usetikzlibrary{intersections, calc}

\usepackage[setpagesize=false]{hyperref}

\usepackage[absolute,overlay]{textpos}
\usepackage{amsmath,amssymb}
\usepackage{amsthm}

\theoremstyle{plain}
\newtheorem{theorem}{Theorem}[section]
\newtheorem{definition}{Definition}[section]

\newtheorem{lemma}{Lemma}[section]
\newtheorem{corollary}{Corollary}[section]

\theoremstyle{definition}
\newtheorem*{Proof}{Proof}
\newtheorem{remark}{Remark}[section]

\numberwithin{equation}{section}

\def\qed{\hfill $\Box$}

\newcommand{\Besov}{B_{p,q}^{s}}
\newcommand{\bunsu}[2]{\frac{#1}{#2}}
\newcommand{\norumu}[2]{\|#1\|_#2} 

\newcommand{\zissu}{\mathbb{R}}
\newcommand{\sizen}{\mathbb{N}}
\newcommand{\seisu}{\mathbb{Z}}
\newcommand{\expect}{\mathbb{E}}
\newcommand{\syugo}[2]{ \{ #1 \mid #2  \}} 

\newcommand{\holder}{\rm{H\ddot{o}lder}}
\newcommand{\argmin}{\mathop{\rm arg~min}\limits}

\allowdisplaybreaks

\begin{document}
\maketitle

\begin{abstract}
Deep learning has achieved notable success in various fields, including image and speech recognition. One of the factors in the successful performance of deep learning is its high feature extraction ability. In this study, we focus on the adaptivity of deep learning; consequently, we treat the variable exponent Besov space, which has a different smoothness depending on the input location $x$. In other words, the difficulty of the estimation is not uniform within the domain. We analyze the general approximation error of the variable exponent Besov space and the approximation and estimation errors of deep learning. We note that the improvement based on adaptivity is remarkable when the region upon which the target function has less smoothness is small and the dimension is large. Moreover, the superiority to linear estimators is shown with respect to the convergence rate of the estimation error. 
\end{abstract}

\section{Introduction}
Machine learning has attracted significant attention, and has been applied to various fields. In particular, deep learning has been in the spotlight owing to its notable success in different fields, for example, image and speech recognition \citep{krizhevsky2012imagenet, hinton2012deep}. Although its success has been confirmed experimentally, the reason why deep learning functions well has not been fully understood theoretically. This problem has been studied in a statistical context by many researchers, and one of the approaches to this problem is the following nonparametric regression problem:
 \[ Y_i=f^\circ (X_i)+\epsilon_i \hspace{0.5cm} (i=1,\ldots ,n ), \]
where $(X_i, Y_i)_{i=1}^{n}$ are observations and $\epsilon_i$ is Gaussian noise. It is usually considered that $f$ is contained in a function class $\mathcal{F}$ and one of the typical criteria used to evaluate an estimator $\hat{f}$ is
\[ \sup_{f^\circ \in \mathcal{F} } \expect[\|f^\circ(X) -  \hat{f}(X)\|_{L_2(P_X)}^2],  \]
where the expectation is taken over the sample observation $(X_i,Y_i)_{i=1}^n$. We call this the worst-case estimation error. We can compare the convergence speed of this quantity among the estimators. In particular, by comparing deep learning with other estimators with respect to this value, the manner in which deep learning works can be analyzed theoretically. Note that analyzing the worst-case error is common in statistics as a theory of minimax-optimality; thus, it has not been specialized to a deep learning setting \citep{tsybakov2008introduction, yang1999information, zhang2002wavelet, khas1979lower, ibragimov1977estimation, stone1982optimal}. We compare the worst-case error for an estimator $\hat{f}$ with the infimum of the worst-case error over all estimators given by 
\[ \inf_{\hat{f} } \sup_{f^\circ \in \mathcal{F} } \expect[\|f^\circ(X) -  \hat{f}(X)\|_{L_2(P_X)}^2],  \]
where the infimum is taken for all measurable mappings that map $n$ observations to $L_2(P_X)$. This is called the minimax optimal risk. 

In previous studies on learning theory, these settings are considered on various function classes, such as a H\"{o}lder space and Besov space. The worst-case error analyses on these function classes have been studied for deep learning as well as several other classic estimators, and their minimax optimal rates have been extensively considered. For example, the rate of a H\"{o}lder space ($C^\beta (\Omega)$, $\Omega \subset \zissu^d$: bounded open domain) is $n^{-\frac{2\beta}{2\beta +d}}$ \citep{stone1982optimal, khas1979lower, ibragimov1977estimation}, whereas that of a Besov space ($B_{p, q}^s ([0, 1]^d)$) is $n^{-\frac{2s}{2s+d}}$\ \citep{kerkyacharian1992density, donoho1996density, donoho1998minimax, gine2016mathematical}. Note that the definitions of a H\"{o}lder space and Besov space are written in the following section. It was shown that deep learning can achieve the near minimax optimal rate. \autoref{tabl1.1} shows the results of previous studies. 

\begin{table}[tb]
  \begin{center}
    \begin{tabular}{ |c||c|c|} \hline
      Function class &  H\"{o}lder( $C^\beta$ ) \  & Besov( $B_{p, q}^s$ )     \\ \hline 
      Approximation error & $ \tilde{O} ( N^{-\frac{\beta}{d}} )\ $ \footnotemark & $\tilde{O } (N^{-\frac{s}{d}})\ $  \\ \hline
      Author & \citet{yarotsky2017error}  & \citet{suzuki2018adaptivity}    \\   \hline
      Estimation error & $\tilde{O}(n^{-\frac{2\beta}{2\beta + d} } (\log n ) ^3 )$ & $\tilde{O}( n^{-\frac{2s}{2s+d}} (\log n )^3 )$ \\ \hline
      Author &  \citet{schmidt2020nonparametric} & \citet{suzuki2018adaptivity}  \\ \hline \hline
      Function class &   variable Besov  ($ B_{p, q}^{s+\beta \| x-c \|_2 ^{\alpha} }$\ )  \\ \cline {1-2}
      Approximation error & $\tilde{O}( N^{-\frac{s}{d}}\ (\log N) ^{-\frac{s- \delta}{\alpha }} )\ $ \\  \cline {1-2}
      Author & This work   \\    \cline {1-2}
      Estimation error & $ \tilde{O} ( n^{-\frac{2s}{2s+d}}(\log n)^{-\frac{2(sd-\nu d-3\alpha s)}{(2s+d)\alpha}} )$ \\  \cline {1-2}
      Author  & This work\\  \cline {1-2}
    \end{tabular}
  \end{center}
  \caption{Approximation and estimation errors of deep learning on function spaces on $[0, 1]^d$. The approximation and estimation errors are measured using the $L_2$-norm and its square, respectively. Here, $N$ is the number of units in each layer of deep neural network and $n$ is the sample size. $s$ and $\beta$ are the parameters of smoothness, and $d$ represents the dimension of the domain where the function spaces are defined. }
  \label{tabl1.1}
\end{table}
\footnotetext{Symbol $\tilde{O}(\cdot)$ indicates the order up to log factors.}

In recent studies, the approximation and generalization capabilities of deep ReLU networks have been studied. In \cite{yarotsky2017error}, the approximation error of deep ReLU networks on a H\"{o}lder space is derived. In addition, \cite{schmidt2020nonparametric} analyzed the estimation error of deep ReLU networks on a H\"{o}lder space and derived the worst-case estimation error that achieves a near optimal rate. \cite{suzuki2018adaptivity} studied the approximation and estimation errors on a Besov space that has a broader function class than a H\"{o}lder class. \cite{suzuki2018adaptivity} noted that deep learning is ``adaptive,'' that is, deep learning can estimate each function effectively by capturing the {\it{local}} smoothness. Indeed, we can see in the analysis by \cite{suzuki2018adaptivity} that adaptivity is important for achieving the near minimax optimal rate. In addition, although deep learning can achieve a near optimal rate even if the spatial homogeneity of the target function smoothness is low, no linear estimator can achieve the optimal rate in such a case. In \cite{pmlr-v89-imaizumi19a}, the superiority was demonstrated when the target functions are piecewise smooth functions. In addition, \cite{hayakawa2020minimax} showed the superiority for a function class with discontinuity and sparsity.  

 In this study, we analyze the generalization capability of ReLU neutral networks on a variable exponent Besov space that has different conditions of smoothness depending on coordinate $x$. Because the smoothness of the target function depends on the input location $x$ (we denote by $s(x)$, i.e., the smoothness of the target function at the location $x$), the difficulty of estimating the function is not spatially uniform. This problem setting highlights the necessity of the adaptivity of the estimators in contrast to previous studies. In previous studies, the estimation problem on this type of function class with non-uniform properties over the input location $x$ has not been analyzed. In addition,  the approximation theory of the variable exponent Besov space has not been studied, although the wavelet decomposition in the variable exponent Besov space has been analyzed, e.g., in \cite{kempka2010atomic}. Therefore, we first need to develop an approximation theory on the variable exponent Besov space using the B-spline basis expansion. In \autoref{general approximation theory}, we derive the lower bound for the general $s(x)$ and analyze the upper bound for the approximation error in the case of $s(x)=s+ \beta\|x - c\|_2 ^\alpha$. In \autoref{deep learning ability}, based on the result in \autoref{general approximation theory}, we derive the upper bounds of the approximation and estimation errors of deep learning. As shown in \autoref{tabl1.1}, the upper bound of the approximation error of deep learning is $\tilde{O}( N^{-\frac{s}{d}}\ (\log N) ^{-\frac{s- \delta}{\alpha }} )$ and that of the estimation error is $\tilde{O} ( n^{-\frac{2s}{2s+d}}(\log n)^{-\frac{2(sd-\nu d-3\alpha s)}{(2s+d)\alpha}} )$, where $N$ is the number of units in each layer of the neural network, $n$ is the sample size, and $\nu$ is a constant, which depends only on $p, d$. The polynomial order of the approximation error depends on the minimum value of $s(x)$, and the poly-log order becomes significant when $\alpha$ is small, that is, the domain around the minimum value of $s(x)$ is small. Moreover, the influence of the poly-log order of the estimation error increases if the dimension $d$ is large and the area around the minimum value of $s(x)$ is small. In \autoref{Superiority to linear estimator}, we show the superiority of deep learning over linear estimators for $0 < p < 2$ or $p=2$ and $0 < \alpha < \frac{d}{3}$: 
 \[ n^{-\frac{2s}{2s+d}}(\log n)^{-\frac{2(sd-\nu d-3\alpha s)}{(2s+d)\alpha}}( \log(\log n) )^{\frac{2d(s-\nu)}{(2s+d)\alpha}} \ll n^{-\frac{2(s-\nu)}{2(s-\nu)+d} }, \] 
where the left-hand side is the worst-case estimation error of deep learning, and the right-hand side is the minimax optimal estimation error over all linear estimators. The contributions of this paper are summarized as follows:
\begin{itemize}
\item For the analysis of the generalization ability of deep learning, we first derive an approximation theory of the variable exponent Besov space using B-spline bases. We derive the lower bound of the approximation error on the variable exponent Besov space with any continuous smoothness function $s(x)$. In addition, we derive the upper bound of the approximation error on the variable exponent Besov space with specific smoothness function $s(x)=s+ \beta\|x - c\|_2 ^\alpha$.

\item To clarify the adaptivity of deep learning, we derive the upper bound of the approximation and estimation errors of deep learning. Subsequently, we show that, as the region where the target function has less smoothness is smaller, the approximation and estimation errors are further improved. This result supports the adaptivity of deep learning. 

\item For a relative evaluation, we compare deep learning with popular linear estimators, such as a least squares estimator, Nadaraya-Watson estimator, and kernel ridge regression. We show the superiority of deep learning over linear estimators with respect to the convergence of the estimation error. 
\end{itemize}

\section{Mathematical preparations}\label{Mathematical preparations}

\subsection{Notation}
In this section, we introduce some of the notations used. 
Throughout this paper, $\Omega$ denotes $[0,1]^d$. For a subset $A$ in $\zissu^d$, we define $\|f\|_{L_r(A)} $ as the $L_r$-norm of a measurable function $f$ on $A$:
\[  \| f \|_{L_r(A)} \coloneqq \left(\int_{A} |f(x)|^r \mathrm{d}x \right)^{\frac{1}{r}}.  \] 
In particular, if $A = \Omega$, we denote $\| \cdot \|_{L_r(A)}$ by $\| \cdot \|_r$. 
Let $(S, \Sigma, P)$ be a pairing of probability space $S$, $\sigma\mbox{-algebra } \Sigma $ on $S$, and probability measure $P$. Furthermore, let $X(s)$ be a random variable on $\zissu^d$ with a probability distribution $P_X$. For a measurable function $f$, we define $\| f \|_{L_r(P_X)}$ as follows: 
\[ \| f \|_{L_r(P_X)} \coloneqq \left( \int_{S} | f(X(s)) |^r {P(\rm{d}s)} \right)^\frac{1}{r}  = \left( \int_{\zissu^d} | f(x) |^r {P^X(\rm{d}\it{x})} \right)^\frac{1}{r}. \] 
Let $X$ be a quasi-normed space. We denote the unit ball of $X$ by $U_X$. That is, \\
\[ U_X \coloneqq \{ x\in X \mid| \|x\|_X \leq 1 \}. \]
We write the support of function $f$ as
\[ \mathrm{supp} f \coloneqq \overline{ \{ x\in \zissu^d \mid f(x)\neq 0 \} }.\]
Let $A \subset \zissu^d$ and $x = (x_1, x_2, \ldots, x_d)^\top \in A$, $p\in \zissu$. In addition, let $a > 0$ and $T$ be a mapping from $\zissu^d $ to $\zissu^d$. Next, we introduce the following notations:
\[ \lceil a \rceil \coloneqq \min\{ n\in \seisu \mid a \leq n  \},  \]
\[ \lfloor a \rfloor \coloneqq \max\{ n\in \seisu \mid a \geq n  \},  \]
\[ \| x \|_p \coloneqq ( x_1 ^p + x_2 ^p +\cdots + x_d ^p )^{\frac{1}{p}}, \]
\[ \mathrm{diam}(A) \coloneqq \sup_{a, b \in A} \| a - b \|_2 , \]
\[ \mathrm{dist}(A, x)\coloneqq \inf_{y \in A} \| y - x \|_2 , \]
\[ B(x, a) \coloneqq \{ y\in \zissu^d \mid \|y - x\|_2 < a\} , \]
\[ T A \coloneqq \{ Ta \mid a \in A \}.\]

\subsection{Nonparametric regression and minimax optimal rate}

In this paper, we consider the following nonparametric regression model:
\[ Y_i=f^\circ (X_i)+\epsilon_i \hspace{0.5cm} (i=1,\ldots ,n ) \tag{1} \label{nonpara},  \]
 where $(X_i, Y_i)_{i=1}^{n}$ is independently identically distributed (i.i.d.) and $X_i \sim P_X$. Here, $P_X$ is a probability distribution on 
 $\Omega$. Moreover, the noise $\epsilon_i$ is i.i.d.
centered Gaussian noise, that is, $\epsilon_i \sim N(0, \sigma^2)\ (\sigma >0)$. We assume that $f^\circ $ is contained in some function class $\mathcal{F}$, that is, $f^\circ \in \mathcal{F}$. We want to estimate function $f^\circ : [0, 1]^d \rightarrow \zissu$ from the observed data $(X_i, Y_i)_{i=1}^{n}$.

For the regression model (\ref{nonpara}), we use the following quantity to evaluate the estimation for each $f^\circ \in \mathcal{F}$:
\[ \expect[\|f^\circ(X) - \hat{f}(X)\|_{L_2(P_X)}^2], \]
where $\hat{f} : [0, 1]^d \rightarrow \zissu$ is the function estimated from the observed data $(X_i, Y_i)_{i=1}^{n}$ and the expectation is taken over the sample observation $(X_i,Y_i)_{i = 1}^n$. To evaluate the quality of estimator $\hat{f}$, we define the following worst-case estimation error: 
\[ R(\hat{f}, \mathcal{F})\coloneqq \sup_{f^\circ \in \mathcal{F} } \expect[\|f^\circ(X) - \hat{f}(X)\|_{L_2(P_X)}^2]. \tag{2} \label{generalize} \]
 Hereafter, we call $R(\hat{f}, \mathcal{F})$ an estimation error for simplicity. We can see from the definition that $R(\hat{f}, \mathcal{F})$ is the worst-case estimation error for $f^\circ \in \mathcal{F}$. We can evaluate the estimators based on the convergence rate of $R(\hat{f}, \mathcal{F})$ as the sample size $n$ increases.

 We use the following lemma to derive the estimation error of deep learning. Note that for a normed space $(V, \| \cdot \|), \mathcal{F}_1 \subset V$ and $\delta > 0$, we denote the $\delta\mathchar`-\mathrm{covering}$ \citep{van1996weak} by $\mathcal{N}(\delta, \mathcal{F}_1, \| \cdot \|_{\infty})$, which represents the minimum number of balls of radius $\delta$ needed to cover $\mathcal{F}_1$.
\begin{lemma}[\cite{schmidt2020nonparametric}, \cite{hayakawa2020minimax}]\label{shmid} {Let $\mathcal{F}_1 $ be a function set and $\hat{f}$ be the least squares estimator in $\mathcal{F}_1$. That is, 
\[ \hat{f}=\argmin_{ f\in\mathcal{F}_1 } \sum_{i=1}^{n} ( Y_i -f(X_i ) )^2. \tag{3} \label{least-square}\] 
Assume that $\| f^{\circ} \|_{\infty} \leq F$ and every element $f\in \mathcal{F}_1 $ satisfies $\| f \|_{\infty} \leq F$ for some $F \geq 1$. If $\mathcal{N}(\delta, \mathcal{F}_1, \| \cdot \|_{\infty} )\geq 3 $ for $\delta > 0$, it then holds that 
\begin{equation*}
\begin{split} \expect[ \| \hat{f} - f^{\circ} \|_{L^2(P_{X})} ]  \leq C &\left[ \inf_{f\in {\mathcal{F}}_1} \| f-f^{\circ} \|_{L^2(P_{X} )}^2 + \right. \\
 &\left.  (F^2 +\sigma^2) \frac{ \log \mathcal{N}(\delta, \mathcal{F}_1, \| \cdot \|_{\infty})}{n} + \delta(F+\sigma) \right].
\end{split}
\end{equation*}
  }
\end{lemma}

In \autoref{shmid}, the first term represents the approximation error, and the second term represents the complexity of the model. To reduce the estimation error, we need to set the complexity of the model such that the first and second terms are balanced. It can be seen from this lemma that we need to derive the approximation error to derive the estimation error. Therefore, we first discuss the approximation error and then derive the estimation error using \autoref{shmid}.

\subsection{Adaptive approximation}
There are two types of approximation methods, non-adaptive and adaptive. For a set of target functions to be approximated, non-adaptive methods fix the basis functions and only change the coefficients of the linear combination. By contrast, adaptive methods change the basis functions and coefficients for each target function. Deep learning is a type of adaptive method because, for each target function, it constructs an appropriate feature extractor that operates as a basis function tailored to each target function. Indeed, deep learning can achieve an (almost) optimal approximation error rate that no non-adaptive method can achieve. Here, we introduce the quantities for an evaluation of these methods and some facts from previous studies.

We define quantities for evaluating both non-adaptive and adaptive methods. We follow the definitions of such quantities presented in \cite{dung2011optimal}.
First, we introduce the quantity used to evaluate a non-adaptive method. Let $X$ be a quasi-normed space defined on a domain $D \subset \zissu^d$ equipped with the norm $\| \cdot\|_X$. Non-adaptive methods are evaluated by the following $N$-term best approximation error\ (Kolmorogov $N$-widths) with respect to the norm of $X$:
\[ d_N (W, X)\coloneqq \inf_{S_N \subset X} \sup_{f \in W} \inf_{g \in S_N} \| f - g \|_X\ , \]
where the infimum is taken over all $N$-dimensional subspaces in $X$. In the definition of $d_N (W, X)$, each function in $W$ is approximated by the fixed $N$ dimensional subspace $S_N$. Thus, the $N$ basis functions are fixed against the choice of $f \in W$, and only the coefficients of linear combinations are changed. 
Next, we introduce two quantities $(\sigma_N, \rho_N)$ that evaluate the adaptive methods. Let $W$ and $B$ be subsets of $X$. The approximation of $W$ by $B$ with respect to the $X$'s norm is evaluated by the following quantity: 
\[ E(W, B, X)\coloneqq \sup_{f\in W} \inf_{\phi \in B} \| f-\phi \|_X.\]
Let $\Phi =\{ \phi_k \}_{k\in \mathcal{K} }$ be a subset of $X$ indexed by a set $\mathcal{K}$\ ($\phi_k \in X$). We define $\Sigma_{N}(\Phi)$ such that it consists of all $N$ linear combinations of the elements of $\Phi$ as 
\[ \Sigma_{N}(\Phi) \coloneqq \left\{ \phi=\sum_{j=1}^{N} a_j \phi_{k_j} : k_j \in \mathcal{K} \right\}.\]
Here, we define the quantity that evaluates the approximation by $N$ linear combinations of functions in $\Phi$ as follows: \[ \sigma_{N} (W, \Phi, X) \coloneqq E( W, \Sigma_{N}(\Phi), X).\]
In contrast to $d_N(W, X)$, for each function in $W$, we can choose $N$ basis functions from $\Phi$ and the coefficients of linear combinations adaptively. Let $\mathcal{B}$ be a family of subsets in $X$. An approximation by $\mathcal{B}$ is evaluated using the following quantity: 
\[ d( W, \mathcal{B}, X ) \coloneqq \inf_{B\in \mathcal{B}} E(W, B, X). \] 
 If $\mathcal{B}$ is the family of all subsets $B$ such that the pseudo-dimension is at most $N$, we denote $d( W, \mathcal{B}, X )$ by $\rho_N (W, X)$, which is called a non-linear $n$-width. Here, the pseudo dimension of $B$ is defined as the largest integer $N$ such that there exist points $a_1, \ldots, a_N \in D$ and $b_1, \ldots, b_N \in \zissu$ that satisfy
\[ |\{ \mathrm{sgn}(y) \mid y=( \mathrm{sgn}( f(a_1)+b_1 ), \ldots, \mathrm{sgn}( f(a_N)+b_N) ), f\in B \}| = 2^N, \]
where $\mathrm{sgn}(x) = \mathbbm{1}_{\{x > 0\}} - \mathbbm{1}_{\{x \leq 0\}}$. 
Because the pseudo-dimension of any $N$-dimensional vector space from a set in $D$ to $\zissu$ is $N$ (see \cite{haussler1992decision}), $\rho_N (W, X)$ 
can evaluate non-adaptive and adaptive methods. 
Note that if $X=L_r(\Omega)$, we denote each quantity by $d_N (W)_r$, $\sigma_{N}(W, \Phi)_r$, and $\rho_N (W)_r$.

It was shown that the approximation error on a Besov space and that on other function spaces related to a Besov space can be improved through an adaptive method. 
We take the example of a Besov space on $\Omega$ and see an improvement when using adaptive methods (see \autoref{def-Besov} for the definition of a Besov space).
First, we determine the approximation error using non-adaptive methods. The lower bound of the non-adaptive methods is 
\[ d_N ( U_{B_{p, q}^{s} })_r    \gtrsim    \begin{cases}
    N^{-\frac{s}{d}+\left(   \frac{1}{p} -\frac{1}{r}  \right) } & ( 1 <  p < r \leq 2, s > d( \frac{1}{p} - \frac{1}{r}  )  ),  \\
    N^{-\frac{s}{d} + \frac{1}{p} - \frac{1}{2} } & ( 1 < p <  2 < r \leq \infty,   s > \frac{d}{p} ),  \\
    N^{-\frac{s}{d}} & ( 2\leq p < r \leq \infty ,   s > \frac{d}{2} ).
  \end{cases} \]
(see \cite{romanyuk2009bilinear, myronyuk2016kolmogorov, vybiral2008widths}). For the functions in $B_{p, q}^{s}$, $s$ controls the smoothness and $p$ controls the spatial homogeneity of the smoothness. In addition, $q$ controls the degree of emphasis on the local smoothness, although it dose not directly influence the convergence rate. However, it was shown in \cite{dung2011optimal} that an adaptive method can achieve the optimal rate of approximation error. Specifically, it was proved in Theorems 5.2 and 5.4 in \cite{dung2011optimal} that, for $0 < p, q, r \leq \infty$ and $d \left(\frac{1}{p} - \frac{1}{r} \right)_{+} < s$, the optimal rate of the approximation methods containing adaptive methods is 
\begin{equation} \label{Dung's inequality}
 \sigma_{N}(U_{B_{p, q}^{s}}, \mathbf{M})_r, \rho_N\ (U_{B_{p, q}^{s} })_r \asymp N^{-\frac{s}{d}}, 
 \end{equation}
where $\mathbf{M}$ is the set of all $M_{k, j}^d$ (see \autoref{Decomposition of functions in Besov space by cardinal B-spline}) whose degree $m$ satisfies $s < \min \{m, m-1 + \frac{1}{p}\}$, which do not vanish identically on $\Omega$. Moreover, an adaptive method can achieve the rate \citep{dung2011optimal}. If parameter $p$ that controls the spatial homogeneity is small, some functions in $B_{p, q}^{s}$ have smooth and rough parts depending on the input location $x$. The usefulness of an adaptive method for $p < 2$ can be interpreted such that adaptive methods can increase the resolution of rough parts, which contributes to an effective approximation. 
 
 In addition, the improvement of the estimation using an adaptive method was shown in \cite{suzuki2018adaptivity} for a Besov space and a mixed-Besov space and in \cite{suzuki2019deep} for an anisotropic-Besov space. By applying an adaptive method to the analysis of the estimation error through deep learning, it was shown that the estimation error could be improved. In \cite{suzuki2018adaptivity} and \cite{suzuki2019deep}, it was proven that the estimation error by deep learning can achieve the minimax rate up to the poly-log order. \\

\subsection{Besov space}

In this section, we introduce the basic properties of the function spaces, in particular, a Besov space and  variable exponent Besov space. First, we define a Besov space as follows:
\begin{definition}[The definition of $B_{p, q}^{s} (\Omega)$]\label{def-Besov}
Let $0< p, q\leq\infty $, $s > 0, r\in \sizen$ and $r > s$. We define the $r$-times difference as
\[
  {\Delta_{h}^r}(f)(x)= \begin{cases}
    \underbrace{\Delta_{h} \circ\Delta_{h} \circ\cdots\circ\Delta_{h}}_{r} (f)(x)  & (x\in \Omega,  x+rh\in \Omega),  \\
    0  & (\rm{otherwise}), \\
  \end{cases}
\]
where $\Delta_{h}(f)(x) = f(x+h)-f(x)$. The $r$-th module of the smoothness is defined as follows:
\[ \omega_{r, p}(f, t)=\sup_{h\in \zissu^d :\norumu{h}{2}\leq t} \| \Delta_h ^r\|_p. \]
We define the following quantity using $ \omega_{r, p}(f, t)$: 
\[
   |f|_{B_{p, q}^{s}(\Omega)}\coloneqq \begin{cases}
    \left[\int_{0}^{1} (t^{-s}\omega_{r, p}(f, t))^{q}\bunsu{1}{t} \mathrm{d}t \right]^{\bunsu{1}{q}} & (q < \infty),  \\
     \sup_{t>0}  t^{-s}\omega_{r, p}(f, t) & (q = \infty). 
  \end{cases}
\]
 By using $|f|_{B_{p, q}^{s}}(\Omega)$, we define the norm as follows:
\[B_{p, q}^{s}(\Omega) = \syugo{f \in {L_p(\Omega)} } {\norumu{f}{{B_{p,q}^{s}(\Omega) }} \coloneqq \norumu{f}{p}+|f|_{\Besov(\Omega)} < \infty}.\]
\end{definition}

\begin{remark}
We note some comments regarding the quantities in \autoref{def-Besov}. 
\begin{itemize}
\item Operator $\Delta_h$ is similar to the differential, and if function $f$ is an $r$-times continuous differentiable function on $\zissu$, it holds that $ \lim_{h \to 0} \frac{ \Delta_h ^r (f)(x) }{|h|^r} = f^{(r)}(x)$.
\item For $\omega_{r, p}(f, t)$, by applying the H\"{o}lder's inequality, it can be easily confirmed that $\omega_{r, p}(f, t)$ becomes larger as $p$ increases. 
\item As $s$ increases, $t^{-s}$ increases for $t$, which is close to $0$. Therefore, when $s$ is larger, $\omega_{r, p}(f, t)$ needs to be smaller for $t$, which is close to $0$. Thus, we can interpret that $s$ controls the local smoothness.
\item If $r \in \sizen$ satisfies $r >s$, the definition of $B_{p, q}^{s}(\Omega)$ does not depend on $r$.
\end{itemize}\end{remark}

It is known that some other function spaces can be reproduced from a Besov space by taking some parameters to satisfy certain conditions. First, we define a $\holder$ space and Sobolev space. 
\begin{definition}[$\holder$ space ($C^\beta (\Omega))$]
Let $\beta > 0$ satisfy $\beta \notin \sizen$. In addition, we define $m\coloneqq\lfloor \beta \rfloor$. For the $m$-times continuous differentiable function $f: \zissu^d \to \zissu$, we define the norm of the $\holder$ space $C^\beta (\Omega)$ as follows:
\[ \| f \|_{C^{\beta}(\Omega)} \coloneqq \max_{|\alpha|\leq m } \| D^\alpha f \|_{\infty} + \max_{|\alpha|=m} \sup_{x, y \in \Omega}
\frac{|D^\alpha f(x) - D^\alpha f(y)| }{|x - y|^{\beta-m}}, \]
where for $\alpha = (\alpha_1, \alpha_2, \ldots, \alpha_d)$, we define $| \alpha |= \sum_{i=1}^{d} | \alpha_i |$, and for $\alpha \in \seisu^d$, we define the derivative by $D^{\alpha} f(x)= \frac{\partial^{|\alpha|} f}{\partial^{\alpha_1}x_1 , \ldots, \partial^{\alpha_d}x_d }$. Using this norm, the $\holder$ space is defined as follows:
\[ C^{\beta}(\Omega) \coloneqq \left\{ f \mid m\  \mbox{times differentiable and}\  \| f \|_{C^{\beta}(\Omega) } < \infty       \right\} .\]
\end{definition}

\begin{definition}[Sobolev space ($W_p ^m (\Omega))$]
Let $m \in \sizen$ and $1\leq p \leq \infty$. For $f \in L_p (\Omega)$, we define the norm of the Sobolev space as follows:
\[ \| f \|_{W_p ^m (\Omega)} \coloneqq \left( \sum_{|\alpha| \leq m} \| D ^{\alpha} f \|_p ^p \right)^{\frac{1}{p}}, \]
where $D ^{\alpha}$ is a weak derivative. Here, $W_p ^m (\Omega)$ is defined as follows:
\[ W_p ^m (\Omega) \coloneqq \left\{ f \in L_p(\Omega) \mid \| f \|_{W_p ^m (\Omega)} < \infty \right\}. \]
\end{definition}
In \cite{triebel1983theory}, the relationships between a Besov space and other function spaces are provided. In addition, the relationships between Besov spaces with different parameters are written. We introduce some of them here. We note that for normed vector spaces $(V_1, \| \cdot \|_{V_1})$ and $(V_2, \| \cdot \|_{V_2})$ such that $V_1 \subset V_2$, if the inclusion map $i : V_1 \to V_2$ is continuous, we denote $V_1 \hookrightarrow  V_2$. 
\begin{itemize}
\item For $m\in \sizen$, $B_{p, 1}^{m} (\Omega) \hookrightarrow W_{p} ^{m} (\Omega) \hookrightarrow B_{p, \infty} ^{m} (\Omega).$
\item$B_{2, 2}^{m} (\Omega) = W_{2}^{m}(\Omega)$.
\item For $0 < s < \infty$ and $s\notin \sizen$,  $C^{s} (\Omega) = B_{\infty, \infty}^{s}(\Omega) $.
\item Let $0 < s < \infty$, $0< p, q, r \leq \infty$ with $s > \delta \coloneqq d\left(\frac{1}{p}-\frac{1}{r}\right)_{+}$. Then, $B_{p, q}^{s}(\Omega) \hookrightarrow B_{r, q}^{s - \delta}(\Omega)$.
\item Let $C_0 (\Omega)$ be the set of continuous functions on $\Omega$. Then, for $s > \frac{d}{p}$, $ B_{p, q}^{s}(\Omega) \hookrightarrow C_0 (\Omega)$.
\end{itemize}

Thus, a Besov space has close relationships between a $\holder$ space and Sobolev space. We can obtain the properties of other function spaces by analyzing the Besov space. 
\subsection{Decomposition of functions in Besov space by cardinal B-spline}\label{Decomposition of functions in Besov space by cardinal B-spline}

For the analysis provided after this section, we introduce some facts regarding the approximation on the Besov space and decomposition using a cardinal B-spline basis researched in \cite{devore1988interpolation} and \cite{dung2011optimal}. In this study, we mainly use the approximation theory of a Besov space with a cardinal B-spline basis, because the authors in \citep{yarotsky2017error, schmidt2020nonparametric} showed the effective approximation for polynomials using a deep ReLU network; thus, a B-spline cardinal basis is convenient for the approximation theory when applying a deep neural network. 

First, we introduce facts regarding $B_{p, q}^s (\Omega)$ studied in \cite{devore1988interpolation}. 
Let $m \in \sizen$ and $N$ be the univariate B-spline basis, i.e., $N(x) \coloneqq \frac{1}{m !} \sum_{j=0}^{m+1} (-1)^j {m+1 \choose j} (x-j)_{+} ^{m}$. In addition, we define the tensor product of the B-splines as
\[ M_{0, 0}^d(x) \coloneqq N(x_1) N(x_2) \cdots N(x_d) \]
and for $k\in \seisu_{+}$ and $ j\in\seisu^d$, we define the cardinal B-spline basis as 
\[ M_{k, j}^d\coloneqq M_{0, 0}^d(2^k(x-j) ) . \] 
$\Lambda(k)$ denotes the set of $j$ in which $M_{k, j} ^d$ does not vanish identically on $\Omega$. Here, it can be shown in the same manner as Corollary 2-2 in \cite{dung2011optimal} that for all $f\in B_{p, q}^s (\Omega)$, there exists $ Q_k (f) = \sum_{j\in \Lambda(k)} c_{k, j} M_{k, j}^d (x)$, which satisfies the following inequality: 
\begin{equation}\label{decomposition inequality} 
 \| f - Q_k (f) \|_r \leq 2^{-k(s-\delta)}, 
\end{equation} 
where $s > \delta = d\left(\frac{1}{p} -\frac{1}{r}\right)_{+}$ and degree $m$ of the cardinal B-spline basis satisfies $s < \min\{m, m-1+\frac{1}{p}\}$. Additionally,  we let $Q_{-1} \coloneqq 0$ and $q_k \coloneqq Q_k -Q_{k-1} $, and it is known that $q_k$ can be represented as follows: 
 \[q_k=\sum_{j\in \Lambda(k)} a_{k, j} M_{k, j}^d (x). \]
 Note that, throughout this paper, $Q_k(f)$ and $q_k(f)$ indicate the decomposition of $f$ above; if the decomposition target $f$ is apparent, we denote these by $Q_k$ and $q_k$. For $s < \min\{m, m-1 +\frac{1}{p}\}$, by Theorem 5.1 and and Corollary 5.3 in \cite{devore1988interpolation}, $f \in B_{p, q}^s(\Omega)$ can be decomposed as follows:
 \begin{equation}\label{tenkai}
 f (x) = \sum_{k=0}^{\infty} \sum_{j\in \Lambda(k)} a_{k, j} M_{k, j}^d . 
 \end{equation}
 Note that the convergence is with respect to the $B_{p, q}^{s}(\Omega)$ norm.  Moreover, the following norm equivalence holds:
  \begin{equation}\label{norm-tenkai}
 \| f\|_{B_{p, q}^{s}(\Omega) } \backsimeq \left( \sum_{k=0}^{\infty} 2^{ksq}\| q_k\|_{p} ^q \right)^{\frac{1}{q}}
 \backsimeq \left( \sum_{k=0}^{\infty} \left( \sum_{j\in \Lambda(k)} | a_{k, j} |^p 2^{(s p-d)k} \right)^{\frac{q}{p}} \right)^{\frac{1}{q}}. 
 \end{equation}

\subsection{Variable exponent Besov space}

Next, we define a variable exponent Besov space. Here, we consider the case in which only parameter $s$ is variable and parameters $p$ and $q$ are fixed. Before the definition of a variable exponent Besov space, we define the log-H\"{o}lder continuity that is assumed for the function of smoothness. 

\begin{definition}[log-H\"{o}der continuity]
For a function $f : \Omega \to \zissu$, if $f$ satisfies the following  log-H\"{o}der continuity, we denote $f\in C_{\rm{log}}(\Omega)$: there exists a constant $c_{\log} > 0$ and 
\begin{equation}\label{logholder}
 | f(x) -f(y) |\leq \frac{c_{\log}}{ \log( e+ \frac{1}{ \| x-y \|_2}) } \quad ( ^\forall x, ^\forall y\in \Omega).
 \end{equation}
\end{definition}
We can see that the log-H\"{o}der continuity is stronger than the continuity, and is weaker than the Lipschitz continuity.

There are some methods to define a variable exponent Besov space. In this study, we define this as follows: 

\begin{definition}[variable exponent Besov space $B_{p, q }^{s(x)}(\Omega)$]\label{variable Besov}
Let $s(\cdot)\in C_{\rm{log}}(\Omega)$.  We assume $0 < \inf_{x\in \Omega} s(x) $ and let $ r\coloneqq \lfloor \sup_{x_\in \Omega } s(x) \rfloor +1$. We define $\omega_{r, p}^{*}(f, t)$ in a similar manner as the case of $B_{p, q}^s(\Omega)$:
\begin{equation}\label{def-omega*}
 \omega_{r, p}^{*}(f, t)=\sup_{h\in \zissu^d :\norumu{h}{2}\leq t} \| t^{-s(\cdot)}\Delta_h^r \|_p . 
\end{equation}
Next, we define $|f|_{B_{p, q}^{s(x)}(\Omega)}$ as follows:
\[
 |f|_{B_{p, q}^{s(x)}(\Omega)}\coloneqq \begin{cases}
 [\int_{0}^{1} (\omega_{r, p}^{*}(f, t))^{q}\bunsu{1}{t} \rm{d}\it{t} ]^{\frac{1}{q}} & (q < \infty), \\
  \sup_{t>0} \omega_{r, p}^{*}(f, t) & (q = \infty). 
 \end{cases}
\]
In the same manner as in $B_{p, q}^{s} (\Omega)$, the norm is defined as the sum of the $L_p$ norm and $|f|_{B_{p, q}^{s(x)}(\Omega)}$, that is, 
\[B_{p, q}^{s(x)}(\Omega) = \syugo{f \in {L_p (\Omega)} } {\norumu{f}{{B_{p,q}^{s(x)}}} \coloneqq \norumu{f}{p}+|f|_{B_{p, q}^{s(x)}} < \infty}.\]

\end{definition}
Note that for the case of $B_{p, q}^s(\Omega)$, $t^{-s}$ is contained in the definition of $|f|_{B_{p, q}^{s}(\Omega)}$; whereas, in the case of a variable exponent, it is contained in the definition of $\omega_{r, p}^{*}(f, t)$. By definition, we can see that the permissible smoothness changes depending on $x$.

We can consider that the definition above is the smoothness parameter of the Besov space $s$ when replaced with variable $s(x)$. From the definition, it holds that 
\[ B_{p, q}^{s(x)}(\Omega) \subset B_{p, q}^{s_{\min}}(\Omega), \]
where $s_{\min} = \min_{x\in \Omega} s(x)$. Note that we denote $\max_{x\in \Omega}s(x)$ by $s_{\max}$ and $\min_{x\in \Omega}s(x)$ by $s_{\min}$. Other similar (probably equivalent) definitions of $B_{p, q}^{s(x)}$ have been studied by \citep{kempka2012spaces, besov2003equivalent}.

Next, we introduce some properties of $B_{p, q}^{s(x)}(\zissu^d)$ that are studied in \cite{almeida2010besov}. 
\begin{itemize}

\item We assume $s_0, s_1 \in L_{\infty}(\zissu^d)$, and for all $x \in \zissu^d$, it holds that $s_0(x) \geq s_1(x)$. We let 
\[ s_0 (x) - \frac{d}{p_0} = s_1(x) - \frac{d}{p_1} + \epsilon(x). \]
If $\inf_{x\in \zissu ^d} \epsilon(x) > 0$ is satisfied, it holds that
\[ B_{p_0, q}^{s_0(x) }(\zissu ^d) \hookrightarrow B_{p_1, q}^{s_1(x)}(\zissu^d).\]

\item For $\delta > 0$, we assume that for all $x \in \zissu^d$, it holds that
\[ s(x) - \frac{d}{p} \geq \delta \max\left\{ 1 -\frac{1}{q},  0 \right\}. \]
 By letting $C_u (\zissu^d)$ be the set of functions that are bounded and uniformly continuous functions, it holds that 
\[ B_{p, q}^{s(x)}(\zissu^d) \hookrightarrow C_{u}(\zissu^d). \]
 
\item We assume that for all $ x \in \zissu^d$, $s(x)$ satisfies $s(x) < 1$. We define the Zygmund space $C^{s(x)}$ as 
\[ C^{s(x)}(\zissu^d) \coloneqq \left\{ f \in L_{\infty}(\zissu^d)\ \middle| \ \| f \|_{\infty} + \sup_{x\in \zissu^d, h\in \zissu^d\backslash \{0\}} \frac{ | \Delta_{h} ^{1} f(x) |}{ | h |^{s(x)} } <\infty \right\}.\]
Then, it holds that 
\[ B_{\infty, \infty}^{s(x)}(\zissu^d) = C^{s(x)}(\zissu^d). \]
\end{itemize}

Thus, it is known that a variable exponent Besov space also reproduces other function spaces by taking certain parameters to satisfy some conditions.

\section{Approximation theory of the variable exponent Besov space and approximation error}\label{general approximation theory}

\subsection{Lower bound of approximation error}

In this section, we evaluate the lower bound of the approximation error on the unit ball of the variable exponent Besov space $U_{B_{p, q}^{s(x)}(\Omega)}$. In the main result (\autoref{lower-bound}), we will show that the polynomial order of the approximation error on $U_{B_{p, q}^{s(x)}(\Omega)}$ cannot be improved from $N^{-\frac{s_{\min}}{d}}$ for any $s(x)$. To prove \autoref{lower-bound}, we show the following lemma. 

\begin{lemma}\label{noreq} 
Let \ $0 < p, q \leq \infty$, $a \in \Omega$, $\xi > 0$ and $\epsilon > 0$. Suppose that s(x) satisfies $\max_{x\in[a-\xi, a+\xi]^d} s(x) < s+\epsilon $ and $f \in U_{B_{p, q}^{s+\epsilon}(\Omega) }$ satisfies $\mathrm{supp} f \subset [ a-\frac{\xi}{2}, a+\frac{\xi}{2} ]^{d}$. Then, there exists $C > 0$ that does not depend on $f$ such that $\| f \|_{B_{p, q}^{s(x)}(\Omega)} \leq C \| f \|_{B_{p, q}^{s+\epsilon}(\Omega)}$ holds. 
\end{lemma}
\begin{Proof}
For $0 < t <\frac{\xi}{2r}$ and $\|h\|_2 \leq t$, $\Delta_{h}^{r} (f)(x)=0$ holds, where $x\in \Omega \backslash [a-\xi, a+\xi]^d $. By (\ref{def-omega*}) it holds that 
\begin{equation}\label{3-1-1}
\omega_{r, p}^{*}(f, t) < t^{-(s+\epsilon)}\omega_{r, p}(f, t). 
\end{equation}
Moreover, by applying a triangle inequality, we have
\[ \| \Delta_{h}^{r} (f) \|_p \lesssim 2^r \| f \|_p. \]
Thus, for any $t \in (0, 1]$, it holds that
\begin{equation}\label{3-1-2}
\omega_{r, p}^{*}(f, t) \lesssim 2^r t^{-s_{\max}} \| f \|_p \leq 2^r t^{-s_{\max}} \| f \|_{B_{p, q}^{s+\epsilon}(\Omega)}, 
\end{equation}
where, for $t \in (0, 1]$, we use $t^{-s(x)} < t^{-s_{\max}}$. Therefore, by (\ref{3-1-1}) and (\ref{3-1-2}), there exists $C_0 > 0$, and we have the following:
\begin{align}
 \int_{0}^{1} (\omega_{r, p}^{*}(f, t) )^q \frac{1}{t} \mathrm{d}t &= \int_{0}^{\frac{\xi}{2r}} (\omega_{r, p}^{*}(f, t) )^q \frac{1}{t} \mathrm{d}t + \int_{\frac{\xi}{2r}}^{1} (\omega_{r, p}^{*}(f, t) )^q \frac{1}{t} \mathrm{d}t \notag\\
 &\leq \int_{0}^{\frac{\xi}{2r}} \{t^{-(s+\epsilon)}\omega_{r, p}(f, t)\}^q \frac{1}{t} \mathrm{d}t + \int_{\frac{\xi}{2r}}^{1} \{2^r \| f \|_{B_{p, q}^{s+\epsilon}(\Omega)} t^{-s_{\max}} \}^q \frac{1}{t} \mathrm{d}t \notag\\
 & \leq {C_0}^q \| f \|_{B_{p, q}^{s+\epsilon}(\Omega)} ^q. \notag
\end{align}
Note that $C_0$ does not depend on $f$, but does depend on $\xi$ and $r$. Therefore, $|f|_{B_{p, q}^{s(x)}(\Omega) }\leq C_0 \| f \|_{B_{p, q}^{s+\epsilon}(\Omega)}$ holds, and we obtain $\| f \|_{B_{p,q}^{s(x)}(\Omega)} \leq (C_0+1) \| f \|_{B_{p, q}^{s+\epsilon}(\Omega)} \equiv C \| f \|_{B_{p, q}^{s+\epsilon}(\Omega)}.$\\
\qed
\end{Proof}

We want to expand the local argument around $s_{\min}$ to the argument on $\Omega$ by using the extension operator. To introduce the extension operator in \autoref{extension}, we define the minimally smooth domain. 

\begin{definition}[minimally smooth domain]
{
Let $S$ be an open set in $\zissu^d$. Here, S is a minimally smooth domain if there exists $\eta > 0$ and open sets $U_{i}\subset \zissu^d\ (i=1, 2, \ldots)$, such that the following conditions hold: \\
(i) For each $x \in \partial S$, ball $B(x, \eta)$ is contained in one of $( U_i)_{i}$, \\
(ii) Point $x \in \zissu^d$ is in at most $N$ sets $U_i$, where $N$ is an absolute constant, and \\
(iii) For each i, $U_i \cap S = U_i \cap S_i$, where $S_i$ is a rotation of a Lipschitz graph domain.\\
Note that $\Lambda$ is a Lipschitz graph domain, if there exists a Lipschitz continuous function $\phi : \zissu^{d-1}\rightarrow \zissu$. That is, there exists $L > 0$, and for all $x, y \in S$, $\frac{|\phi(x) -\phi(y)|}{\| x-y \|_2} \leq L$ holds, and $\Lambda$ can be written as follows: 
\[\Lambda =\{(u, v) :   u\in \zissu^{d-1}, v\in \zissu  \ \text{and}\  v > \phi(u) \} . \]
Here, $(u, v)$ indicates that $(u^\top, v)^\top$ for $ u\in \zissu^{d-1}$ and $ v\in \zissu$. 
}
\label{minimally smooth}
\end{definition}
 In \cite{stein1970singular}, it is stated that all convex sets in $\zissu^d$ are minimally smooth domains. For a better understanding of \autoref{minimally smooth}, we prove that a cube is a minimally smooth domain. 
\begin{lemma}\label{cube-minimal}{Let $A$ be the interior of $[0,1]^d$. Then, $A$ is a minimally smooth domain.}
\begin{Proof}
The value of $x \in \partial A$ can be expressed as follows: 
\begin{align*}
x &= \sum_{i=1}^{d-1} t_i w_{j_i} + t_{j_d}w_{j_d}\label{*}\\
(1\leq j_1 < j_2 < &\ldots < j_{d-1}\leq d , j_{d} = \{1, \cdots, d\} \backslash \{j_i\}_{i=1}^{d-1}, t_{j_d} \in \{ 0, 1\}) \notag , 
\end{align*}
where $\{w_i\}_{i=1}^{d}$ is a standard basis in $\zissu^d$ and $0 \leq t_i \leq 1\ ( i=1,\ldots, d-1)$.

Let $A^{'}_{(0, \ldots, 0)}$ be the image of $A$ under an orthonormal transformation $O_{(0, \ldots, 0)}$ that transfers $(1, 1, \ldots, 1)$ to $(0, 0, \ldots, \sqrt{d})$. We also define 
\[ D \coloneqq \left\{ \sum_{i=1}^{d} t_i w_i \ \middle|\ 0\leq t_i \leq \frac{2}{3}\right\}. \] 
Here, we define $h: \zissu^{d-1} \rightarrow \zissu$ as follows:
\[ h(u) \coloneqq 
\begin{cases}
\inf_{(u, v)\in \partial A^{'}_{(0, \ldots, 0)}} v &(\text{if there exists}\ v \in \zissu \ \text{that satisfies}\ \\
\ &\ (u, v)\in \partial A^{'}_{(0, \ldots, 0)} ), \\
\inf_{(u^{'}, v)\in \partial A^{'}_{(0, \ldots, 0)}} v + \|u - u^{'} \|_2 &(\text{otherwise}), 
 \end{cases}
 \]
where $u^{'} = \argmin_{^\exists v \in \zissu, s.t., (u^{'}, v) \in \partial A^{'}_{(0, \ldots, 0)} } \| u - u^{'} \|_2$. It is clear that $h(u)$ is a Lipschitz continuous function. We also let $S_{(0, \ldots, 0)} $ be the following: 
\[S_{(0, \ldots, 0)} \coloneqq \{(u, v) : u\in \zissu^{d-1}, v\in \zissu \ \text{and} \ v > h(u) \}. \]
For each point $x$ in $D$, based on the definition of $D$ and radius $\frac{1}{3\sqrt{d}}$, it holds that 
\begin{equation*}
 B\left( x, \frac{1}{3\sqrt{d}} \right) \cap A = B\left( x, \frac{1}{3\sqrt{d}} \right)\cap O_{(0, \ldots, 0)}^{-1} S_{(0, \ldots, 0).} 
\end{equation*}
We define the vertex set of $[0, 1]^d$ as $\{ z_1, z_2, \ldots, z_{2^d} \} \subset \zissu^d$. For each vertex $z_j$, through the translation that transforms $z_j$ into $(0, \ldots, 0)$, we can apply the same argument as $(0, \ldots, 0)$ and obtain $S_{z_i}$; it thus holds that 
\begin{equation}\label{defi(iii)}
 B\left( x, \frac{1}{3\sqrt{d}} \right) \cap A = B\left( x, \frac{1}{3\sqrt{d}} \right) \cap O_{z_j}^{-1}S_{z_j}, 
\end{equation}
where $x$ is in the neighborhood of $z_j$, which corresponds to $D$. We set $\ell \in \sizen$ such that it satisfies $\frac{1}{2^{\ell}} < \frac{1}{6\sqrt{d}}$. For all $x\in \zissu^d$ with each coordinate $x_i = \frac{j}{2^{\ell}} \ (1\leq i \leq d, 0\leq j \leq2^{\ell} )$, we number $x$ of index $k$ as $x(k)$ and denote the set of $B(x(k), \frac{1}{3\sqrt{d}})$ as $\{ U_k \}_{k=1, \ldots, n}$. If we take $0 < \eta < \frac{1}{6\sqrt{d}}$, condition (i) in \autoref{minimally smooth} is satisfied. Condition (ii) in \autoref{minimally smooth} is also satisfied, and condition (iii) is satisfied by (\ref{defi(iii)}).

\qed
\end{Proof}
\end{lemma}

\begin{lemma}\label{extension} {Let $S \subset \zissu^d$ be a closed subset whose interior is a minimally smooth domain. Then, for $0 < p, q \leq \infty$ and $0 < s$, there exists an extension operator $\mathscr{E}: f\in B_{p, q}^{s}(S) \mapsto \mathscr{E}f \in B_{p, q}^{s} (\zissu^d)$ such that 
\begin{itemize}
\item $\mathscr{E}$ is a bounded mapping, that is, there exists a constant $C$ that does not depend on $f \in B_{p, q}^{s}(\Omega)$, and $ \| \mathscr{E}f \|_{B_{p, q}^{s}(\zissu^d) }\leq C\| f \|_{B_{p, q}^{s} (S) }$ holds,
\item $\mathscr{E}f(x) =0$, where $\mathrm{diam}(S)=\xi$ and $\mathrm{dist}(S, x) > 6\xi$.
\end{itemize}
 }
\begin{Proof}
Although \cite{devore1993besov} proved this for only $0 < p \leq1$, it can also be proved for $1\leq p < \infty$ using the technique in Theorem 6.6 in \cite{devore1993besov}. Furthermore, for $p = \infty$, 
it can be proven in the same way as the case $0 < p \leq1$. \\
\qed
\end{Proof}
\end{lemma}
By using \autoref{noreq}, \autoref{cube-minimal}, and \autoref{extension}, \autoref{lower-bound} can be proven. We redefine $\mathbf{M}$ as the set of all $M_{k, j}^d$, whose degree $m$ satisfies $s_{\max} < \min \{m, m-1 + \frac{1}{p}\}$.

\begin{theorem}\label{lower-bound}{Suppose $s_{\min} > d\left(\frac{1}{p} - \frac{1}{r} \right)_{+}$. Then, for all $\epsilon>0$,  it holds that 
\[\sigma_N \left( U_{B_{p, q}^{s(x)} }, \mathbf{M}\right)_r, \rho_N \left( U_{B_{p,q}^{s(x)} } \right)_r \gtrsim N^{-\frac{s_{\min}+\epsilon}{d} }.\]}
\begin{Proof}
Let $s(a) = s_{\min}$. By the continuity of $s(x)$, 
 for all $\epsilon > 0$, there exists $\xi > 0$ such that $| x- a | < \xi \Rightarrow | s(x)-s_{\min} | < \epsilon$.
Let $Q \coloneqq [a-\frac{\xi}{14\sqrt{d}}, a+\frac{\xi}{14\sqrt{d}}]^d$.
By \autoref{cube-minimal}, $Q$ is a minimally smooth domain. Thus, by applying \autoref{extension}, there exists
 $\mathscr{E}f : B_{p ,q} ^{s_{\min}+\epsilon}(Q) \to B_{p, q}^{s_{\min}+\epsilon}(\Omega)$ 
that satisfies the two properties in \autoref{extension}. Note that \autoref{extension} can also be used for extending functions to $\Omega$ because for a function, $g : \zissu^d \to \zissu$, it holds that $\| g \|_{B_{p, q}^{s}(\Omega) } \leq \| g \|_{B_{p, q}^{s}(\zissu^d) }$. 
By using these two properties and \autoref{noreq}, it can be proven that the image of the restriction mapping $f\in U_{B_{p, q}^{s(x)}(\Omega)} \mapsto f|_Q $ contains $ \syugo{f\in B_{p, q}^{s_{\min}+\epsilon} (Q) }{\| f \|_{B_{p, q}^{s_{\min +\epsilon}}(Q)} \leq \frac{1}{C}}$, where $C > 0$ is a constant.\\
 Let $Ef$ be an approximation function of $f$. Here, the following inequality clearly holds: 
 \[ \| f-Ef \|_{L^r(\Omega)} \geq \| f-Ef \|_{L^r(Q)}. \]
Note that the set of functions that satisfy the condition with respect to $\rho_N$ on $\Omega$ are mapped subjective to the set of functions that satisfy the condition on $Q$ by the restriction. By applying (\ref{Dung's inequality}), under each condition $\sigma_N$ or $\epsilon_N$, the following inequality holds: 
\begin{align}
 \inf_{Ef} \sup_{f \in B_{p, q}^{s(x)} (\Omega)} \| f-Ef \|_{L^r(\Omega)}
 &\gtrsim \inf_{Ef} \sup_{f \in B_{p, q}^{s_{\min}+\epsilon} (Q)} \| f-Ef \|_{L^r(Q)} \notag\\
 &\asymp N^{-\frac{s_{\min}+\epsilon}{d}}. \notag
 \end{align}
Thus, the proof is completed.\\
\qed
\end{Proof}
\end{theorem}

\begin{remark}
\autoref{lower-bound} indicates that even an adaptive method cannot improve the polynomical order of an approximation error better than $N^{-\frac{s+\epsilon}{d} }$ for any $\epsilon > 0$. However, it should be noted that the constant $C_{\epsilon}$ hidden in $\gtrsim$ satisfies $\lim_{\epsilon \to 0} C_{\epsilon} = \infty$. In contrast, by the inclusion relation $ B_{p, q}^{s(x)}(\Omega) \subset B_{p, q}^{s_{\min}}(\Omega) $, it is known that $N^{-\frac{s_{\min} }{d} }$ can be achieved \citep{dung2011optimal}. Therefore, our interest is in improving the rate by a factor slower than a polynomial order. In fact, we will show that a poly-log order improvement can be realized. 
\end{remark}

\subsection{Upper bound of approximation error}

In this section, we analyze the approximation error using the adaptive method. Because it is difficult to deal with a general $s(x)$, we analyze a specific $s(x)$ defined by 
\[ s(x) = s+\beta\| x - c \|_2 ^\alpha \quad ( \alpha, \beta, s > 0, c\in \Omega). \]
Throughout this paper, we fix parameters $\alpha, \beta, s > 0, c\in \Omega$. It is clear that $s(x)$ takes the minimum value at $x = c$, and as $\alpha$ decreases, the gradient of $s(x)$ around $c$ becomes larger. We note in our analysis that the gradient of $s(x)$ around the minimum point is important for the order of the approximation error. Thus, this form of $s(x)$ is convenient and sufficient to characterize the approximation of the variable exponent Besov space. 
Consider the case in which $s(x)$ takes the minimum value at several points, and the gradient of $s(x)$ at each minimum point behaves as in a single minimum situation. The approximation error of this case is the same as that of $s(x)$, taking the minimum value at only one point. For example, for $d = 1$, $s(x) = 1+ \sqrt{|x-\frac{1}{4}| }\mathbbm{1}_{[0, \frac{1}{2}]} (x)+ \sqrt{|x-\frac{3}{4}| }\mathbbm{1}_{[\frac{1}{2}, 1]} (x)$ falls in our analysis.

\begin{figure}
  \begin{center}
   \includegraphics[width=70mm]{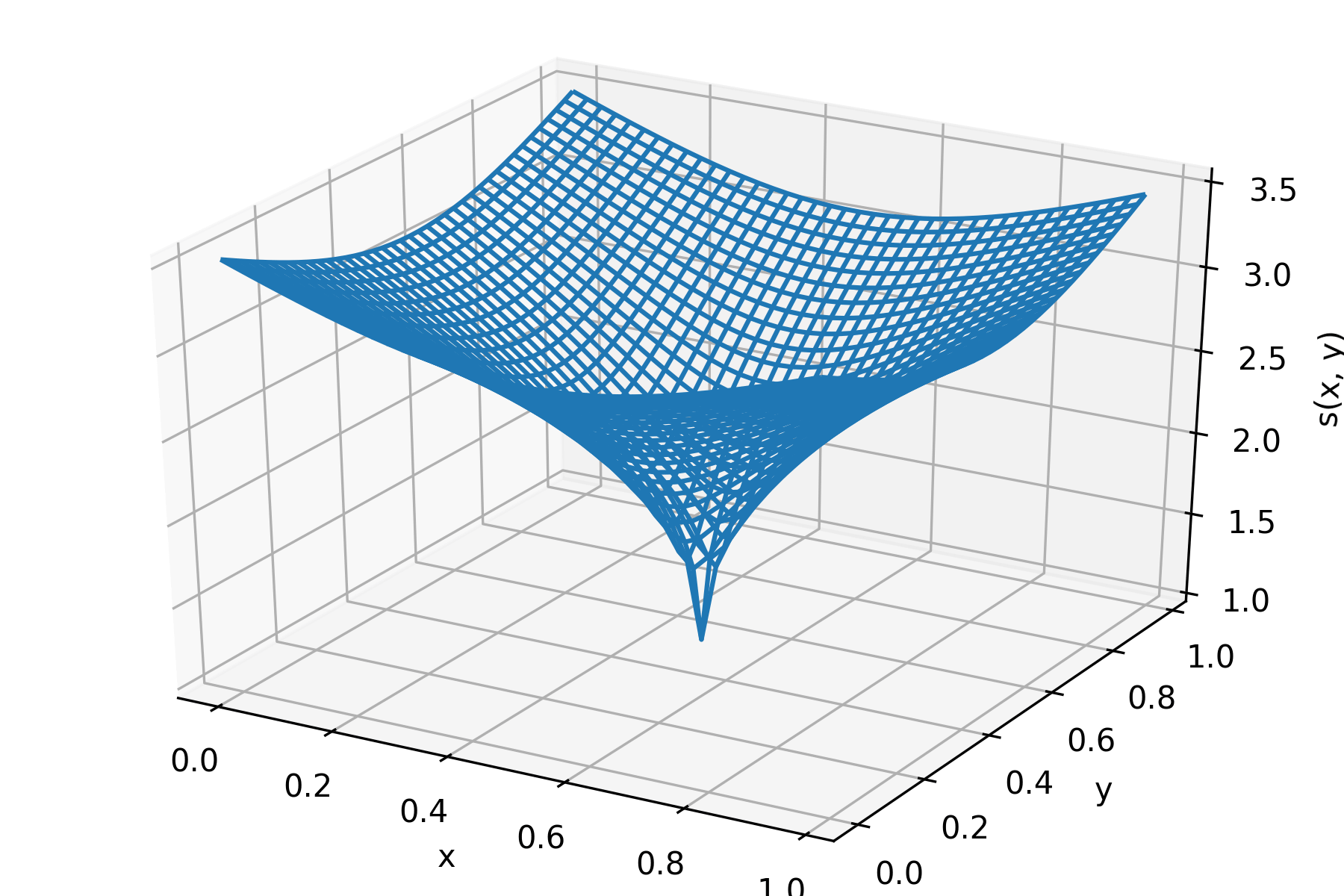}
  \end{center}
  \caption{Example in $\zissu^2$: $s(x)=1+3 \sqrt[4] {(x- \frac{1}{2})^2 + (y- \frac{1}{2})^2 } $.}
  \label{3Ds(x)}
 \end{figure}
Let us determine whether $s(x)$ satisfies the log-$\holder$ continuity. 
If $\alpha > 1$, it is clear that $s(x)$ satisfies the log-$\holder$ continuity because $s(x)$ is a Lipschitz continuous function. Thus, it suffices to consider the case of $0 < \alpha \leq 1$. Because if we exclude the neighborhood of $c$, $s(x)$ is a Lipschitz continuous function in $\Omega$, it suffices to consider a neighborhood of $c$. Based on the triangle inequality and $0 <\alpha \leq 1$, the following inequality holds:
 \[ \| x - c \|_2 \leq \| y - c \|_2 + \| x - y \|_2 \leq \left( \| y - c \|_2 ^\alpha + \| x - y \|_2 ^\alpha \right)^{ \frac{1}{\alpha} }. \]
 Thus, \[ \quad \| x - c \|_2 ^\alpha - \| y - c \|_2 ^\alpha \leq \| x - y\|_2 ^\alpha. \]
Therefore, it holds that
\begin{align}
| s(x) - s(y) | \log \left( e+ \frac{1}{ \| x-y \|_2 } \right) & = \beta | \| x - c \|_2 ^\alpha - \| y - c \|_2 ^\alpha | \log \left( e + \frac{1}{ \| x - y \|_2 } \right) \notag \\
 & \leq \beta \| x - y \|_2 ^\alpha \log \left( e + \frac{1}{ \| x - y \|_2 } \right). \notag
 \end{align}
By $\lim_{t \to \infty} \frac{ \log(e + t) }{t^\alpha} = 0$, the log-$\holder$ continuity is confirmed. \\

Let $q_k=\sum_{j\in \Lambda(k)} a_{k, j} M_{k, j}^d (x)$. Here, let $A$ be a subset in $\zissu^d$, and denote the set of indexes $j$ that satisfy $\mathrm{supp} M_{k, j} ^d \cap A \neq \emptyset$ by $\Lambda_{A} (k)$. Moreover, let $ m_{A, k} \coloneqq | \Lambda_A (k) |$. We reorder indexes $j \in \Lambda_A (k)$ as $\{v_{A, j} \}_{j=1}^{m_{A, k} }$ such that the coefficients are ordered in descending order. That is, the following inequality holds: 
\[ | a_{k, v_{A, 1} } | \geq | a_{k, v_{A, 2} } | \geq \cdots \geq | a_{k, v_{m_{A,k}} } |. \]
Here, we denote
\[ G_k(q_k, m, A) \coloneqq \sum_{j=1}^{m} a_{k, v_{A, j} } M_{k, v_{A, j} } ^d. \]
 \\
Subsequently, we let $\delta \coloneqq d\left(\frac{1}{p} - \frac{1}{r}\right)_{+}$.\\ 
To prove the \autoref{upper-bound} that provides an upper bound of the approximation error in the case of $s(x)= s+\beta\| x - c \|_2 ^\alpha$, let us show \autoref{sup-th2}.

\begin{lemma}\label{sup-th2}
{
Let $0 < p, q, r \leq \infty$, $t > 0$, $ A=[c-t, c+t]^d$, and $s(x)= s+\beta\| x - c \|^\alpha$. Suppose that $s > \delta$ and the degree of cardinal B-spline $m$ satisfy $s_{\max} < \min\{m, m-1+\frac{1}{p} \}$. Moreover, we let $\lambda > 0$, $0 < \epsilon < \frac{d(s -\delta) }{\delta} $, $N \asymp 2^{\bar{k}d} $, $k^{*} = \lceil \epsilon^{-1} \log(\lambda 2^{\bar{k}d} )\rceil+\bar{k}, (\bar{k}+N_k)^{*} = \lceil\epsilon^{-1} \log (\lambda 2^{\bar{k}d}) \rceil + \bar{k} +N_k , n_k = \lceil \lambda 2^{\bar{k}d} 2^{-\epsilon(k-\bar{k} )}\rceil\ and\ m_k = \lceil \lambda 2^{\bar{k}d} 2^{-\epsilon(k-\bar{k}-N_k )} \rceil$. For $f\in U_{B_{p,q}^{s(x)}(\Omega) }$, we define $f_N$ as follows:\\
(i) Suppose that $p\geq r$,
\[ f_N = Q_{\bar{k}}( f ) \mathbbm{1}_{A ^c} + Q_{\bar{k}+N_k }(f) \mathbbm{1}_{A},\]
(ii) Suppose that $p < r$, 
\begin{align}
 f_N &=Q_{\bar{k}}(f) \mathbbm{1}_{A^c} + \sum_{k=\bar{k}+1}^{k^*} G_k( q_k, n_k, A^c ) \mathbbm{1}_{A^c} + Q_{\bar{k}+N_k }(f) \mathbbm{1}_{A} \notag\\
 &+ \sum_{k=\bar{k} + N_k +1}^{(\bar{k}+N_k)^{*}} G_k( q_k, m_k, A ) \mathbbm{1}_{A} \notag \\
 &= Q_{\bar{k}}(f) \mathbbm{1}_{A^c} + \sum_{k=\bar{k}+1}^{k^*} \sum_{j=1}^{n_k} a_{k, v_{A^c, j} } M_{k, v_{A^c, j} }^{d} \mathbbm{1}_{A^c} + Q_{\bar{k}+N_k }(f) \mathbbm{1}_{A} \notag\\
 &+ \sum_{k=\bar{k} + N_k +1}^{(\bar{k}+N_k)^{*}} \sum_{j=1}^{m_k} a_{k, v_{A, j} } M_{k, v_{A, j} }^{d} \mathbbm{1}_{A}, \notag
\end{align}
where $Q_{k}(\cdot)$ is the approximation determined through a B-spline basis for the functions in $B_{p, q}^{s}(\Omega)$ defined in \autoref{Decomposition of functions in Besov space by cardinal B-spline}. Under this condition, the inequality below holds:
\[ \| f - f_N \|_r \lesssim
\begin{cases}
2^{-\bar{k}(s+\beta t^{\alpha} ) } +2^{-(\bar{k}+N_k)s} & ( p \geq r ), \\
2^{-\bar{k}(s+\beta t^{\alpha} ) } + 2^{- s \bar{k}} 2^{- (s-\delta)N_k} & ( p < r ).
\end{cases}
\]
}

\end{lemma}
The proof is given in Appendix \ref{proof-of-sup}.

\begin{theorem}\label{upper-bound}{Let $s(x) = s+ \beta \| x - c \|_2 ^\alpha $ and suppose $0 < p, q, r \leq \infty$ and $s > \delta$. For all $f\in U_{B_{p,q}^{s(x)}(\Omega) }$, there exists $f_N$ that is represented as an N linear combination of cardinal B-spline times some indicator function and satisfies the following inequality: $$\| f- f_N \|_{r} \lesssim N^{-\frac{s}{d}} \left( \frac{ \log N }{\log(\log N)} \right)^{-\frac{s- \delta}{\alpha }}.$$ }
\begin{Proof}
Let the degree $m$ of cardinal B-spline satisfy $s_{max} < m$. Note that $f\in B_{p,q}^{s(x)} \Rightarrow f\in B_{p, q}^{s}$.
In addition, let $\bar{k}\in \sizen$ satisfy $N\asymp 2^{\bar{k}d}$. 

We take the adaptive approximation method around the minimum of $s(x)$. Let $a_k$ be a positive number depending on $N$, which will be fixed below. It is clear that for $a_{k} > 0$,
\[ { 2^ {\bar{k}(s+\beta t^{\alpha}) } \geq a_{k}2^{\bar{k}s } } \Rightarrow t \geq \left( \frac{1}{\beta} \right)^{\frac{1}{\alpha}} \left( \frac{\log a_{k}}{\bar{k}} \right)^{\frac{1}{\alpha}}.\]
Let $t \coloneqq \left( \frac{1}{\beta} \right)^{\frac{1}{\alpha}} \left( \frac{\log a_{k}}{\bar{k}} \right)^{\frac{1}{\alpha}}$, and let $A \coloneqq [c-t, c+t]^d$. We consider increasing the resolution of B-spline in $A$, such that the number of B-splines on $A$ satisfies $\asymp 2^{\bar{k}d}$. 
That is, the following formula holds: 
\[2^{(\bar{k}+N_k)d} \left( \frac{\log a_{k}}{\bar{k}} \right)^{\frac{d}{\alpha}} \asymp 2^{\bar{k}d}. \] Thus, $N_k$ is defined as $ N_k= \lceil \log \left( \frac{\bar{k}}{\log a_{k}} \right) ^{\frac{1}{\alpha}} \rceil$. We have the approximation error by \autoref{sup-th2}, 
\[ \| f - f_N \|_r \lesssim
\begin{cases}
2^{-\bar{k}(s+\beta t^{\alpha} ) } +2^{-(\bar{k}+N_k)s} & ( p \geq r ), \\
2^{-\bar{k}(s+\beta t^{\alpha} ) } + 2^{- s \bar{k}} 2^{- (s-\delta)N_k} & ( p < r ).
\end{cases}
\]
 Therefore, it holds that
\[ \| f -f_N \|_r \lesssim 2^{-\bar{k}s } \left(\frac{1}{a_{k}} + \left( \frac{\bar{k}}{\log a_{k} }\right) ^{-\frac{s-\delta}{\alpha }} \right).\]
Here, setting $a_{k} = \left(\frac{\bar{k}}{\log \bar{k}} \right)^{\frac{s-\delta}{\alpha }}$, we then obtain the desired result.
\qed
\end{Proof}
\end{theorem}

\begin{remark}
Here, $\beta$ does not appear in the convergence rate in the \autoref{upper-bound}, but is hidden in the constant term. In addition, note that the constant term in $\lesssim$ is taken independent of $c$. 
\end{remark}
　
\begin{remark}[approximation theory for general $s(x)$]
The method in \autoref{upper-bound} can be applied to a general $s(x)$. Note that, if the measure of $x$ satisfying $s(x)=s_{\min}$ is not $0$, the method does not make sense, that is, the approximation error is not better than $N^{-\frac{s_{\min}}{d}}$ including a poly-log order. A summary of the approximation method is as follows:
\begin{align}
1.\ &\mbox{Set a positive integer}\ a_k, \mbox{and increase the resolution of the domain}\notag \\ &A=\left\{ x\in \Omega \mid 2^{\bar{k} s(x)}\leq a_k 2^{\bar{k}s_{\min}} \right\}\ \mbox{as}\ 2^{(\bar{k}+N_k)d} \mu(A)\asymp 2^{\bar{k}d}, \notag \ \mbox{where}\ 2^{\bar{k}d} \asymp N , \\
2.\ & \mbox{Fix}\ a_k\ \mbox{so that}\ 2^{\bar{k}s_{\min}} a_k \asymp 2^{({\bar{k}+N_k})s_{\min}}. \notag
\end{align}
 It can be seen that the measure around the minimum point is important for the approximation error rate of this method. This is controlled by exponent $\alpha$ in $s(x)=s+\beta\| x - c \|_2 ^\alpha$. This also indicates that our analysis for the specific choice of $s(x)$ provides essential insight for more general situations.
 \end{remark}
It can be seen that the poly-log part of the order $ N^{-\frac{s}{d}} \left( \frac{ \log N }{\log(\log N)} \right)^{-\frac{s- \delta}{\alpha }} $ descreases as $\alpha$ decreases. That is, as the gradient of $s(x)$ around the minimum point sharpens, the approximation error improves. Moreover, for $r < p$, the poly-log order does not depend on dimension $d$. Thus, if $d$ is large, the poly-log factor has a relatively strong effect. The dependence of the poly-log order on $p$ can be interpreted as follows. Because $p$ controls the homogeneity of the smoothness of functions in a variable exponent Besov space, if $p$ is small, the number of B-spline bases for the adaptive method around the minimum of $s(x)$ should be larger.

The adaptive method in \autoref{upper-bound} is taken at around $c$. Note that, if $c$ is fixed and $r \leq p$, the B-spline bases can be fixed in a non-adaptive manner, that is, we may fix the bases independent of the target function $f$. Therefore, the corollary below immediately follows. 
\begin{corollary}{Let $s(x)=s+ \beta \| x - c \|_2 ^\alpha $ and suppose $0 < q, r \leq \infty$, $p \geq r$. Then, \[d_{N}\left( U_{B_{p, q}^{s(x)}(\Omega)} \right)_r \lesssim N^{-\frac{s}{d}} \left( \frac{ \log N }{\log(\log N)} \right)^{-\frac{s}{\alpha }}. \]
}
\begin{Proof}
In the proof of \autoref{upper-bound}, $f \in B_{p, q}^{s+\beta\| x - c \|_2 ^\alpha}(\Omega)$ can be approximated by a fixed N linear combination of the B-spline basis times an indicator function. 
\qed
\end{Proof}
\end{corollary}
\begin{remark}
Here, we evaluate how effective the adaptive approximation is to improve the accuracy. We compare the approximation error of $B_{p, q}^{s+\beta\| x - c \|_2 ^\alpha}(\Omega)$ with that of $B_{p, q}^{s+\epsilon}(\Omega)$. 
By calculating the upper bound of $\epsilon$ that satisfies the following inequality, 
\[ N^{-\frac{s+\epsilon}{d} } \leq N^{-\frac{s}{d}} \left( \frac{ \log N }{\log(\log N)} \right)^{-\frac{s-\delta}{\alpha }},\]
we can see that the approximation error by an adaptive method is equivalent to that of $B_{p, q}^{s+\epsilon}(\Omega)$, where $\epsilon = \frac{ \log \left(\frac{ \log N }{\log (\log N)} \right)^{\frac{(s-\delta)d}{\alpha}} }{\log N}$. Therefore, it can be seen that for $N$, which is not too large, the improvement using the adaptive method is significant if $\alpha$ is small. 
\end{remark}

\section{Approximation and estimation errors of deep learning}\label{deep learning ability}

\subsection{Approximation error of deep learning}
In this section, we evaluate the approximation and estimation errors of  deep neural networks on $B_{p, q}^{s+\beta\| x - c \|_2 ^\alpha}(\Omega)$. We denote the ReLU activation function by $\eta(x) = \max \{x, 0\}$, where $\eta(x)$ is operated in an element-wise manner for vector $x$. We define the neural network with the ReLU activation, depth $L$, width $W$, sparsity constraint $S$, and norm constant $B$ as follows:
\begin{align}
&\Phi( L, W, S, B ) \coloneqq \{ ( A^{(L)} \eta( \cdot ) + b^{(L)} ) \circ \cdots \circ ( A^{(2)} \eta( \cdot ) + b^{(2)} )  \circ ( A^{(1)}x + b^{(1)} )\mid \notag \\
&A^{(1)} \in \zissu^{W\times d}, A^{(L)} \in \zissu^{1\times W}, A^{(k)} \in \zissu^{W\times W} \quad( 1< k < L), \notag\\
&b^{(L)} \in \zissu, b^{(k)} \in \zissu^{W} \quad(1 \leq k < L ), \notag\\
&\sum_{j=1}^{L} ( \| A^{(j)} \|_0 + \| b^{(j)} \|_0 ) \leq S , \quad \max_{j} \{ \| A^{(j)} \|_{\infty} \vee \| b^{(j)} \|_{\infty} \} \leq B\}, \notag 
\end{align}
where, for matrix $A$, $\| A \|_{\infty}$ is the maximum absolute value of $A$ and $\| A\|_{0}$ is the number of non-zero elements of $A$. 

We evaluate the worst-case approximation error of the deep neural networks $\Phi( L, W, S, B )$ on $U_{B_{p, q}^{s+\beta{\|x-c \|_2}^\alpha}}(\Omega)$  with  $L_r$ norm. That is, the quantity we want to evaluate is
\[ \sup_{f^\circ \in U_{B_{p, q}^{s+\beta\| x-c \|_2 ^\alpha}(\Omega) }} \inf_{\hat{f} \in \Phi(L, W, S, B)} \| f^\circ - \hat{f} \|_r.\]

We assume that there exists a density function of $P_X$ with respect to the Lebesgue measure that is denoted  $p(x)$. Moreover, we assume that there exists $T > 0$ such that $p(x) \leq T$ for all $x \in \Omega$. 

First, we use the lemma below. This indicates the approximation of the B-spline function by a neural network.
\begin{lemma}[\cite{suzuki2018adaptivity}]\label{B-spappro}{Let\ $m \in\sizen$\ be the degree of\ $M_{0, 0}^d$ and \ let \ $c_{ (d, m) }$ be a constant that depends only on \ $d, m$. For all $\epsilon > 0$, there exists a neural network \ $\bar{M}\in \Phi(L_0, W_0, S_0, B_0)$ with $L_0 \coloneqq 3+2\lceil \log_2 (\frac{3^{d\vee m}}{ \epsilon c_{(d, m)} }) +5 \rceil \lceil \log_2 (d\vee m) \rceil, W_0 \coloneqq 6dm(m+2)+2d, S_0\coloneqq L_0 {W_0}^2$\ and\ $ B_0 \coloneqq 2(m+1)^m$ that satisfies
\[ \|M_{0, 0}^d - \bar{M} \|_{L^{\infty}(\zissu^d)} \leq \epsilon, \] and for all $x \notin [0, m+1]^d, \bar{M}(x)=0$.}
\end{lemma}

\autoref{approxi} indicates that the upper bound of the approximation error of $U_{B_{p, q}^{s+\beta\| x - c \|_2 ^\alpha}(\Omega)}$ by a neural network. The proof is similar to that of Proposition 1 in \cite{suzuki2018adaptivity}. The outline of the proof aims to approximate $f_N$ in \autoref{upper-bound} by a neural network. The proof is presented in Appendix \ref{proof-approxi}. 

\begin{theorem}\label{approxi}{Suppose that $0< r < \infty$, $ 0 < p, q, \leq \infty, 0 < s <\infty$, and $s(x)=s+\beta \| x - c \|_{2} ^{\alpha}$}. Furthermore, we assume that $s > \delta$ and 
$m \in \sizen$ satisfies $ s_{\max} < \min\{ m, m-1+\frac{1}{p} \} $. Let $\nu \in \zissu$ be $\nu =\frac{1}{2} \min\{ \frac{d(s -\delta)}{\delta}, 1 \}$. For a sufficiently large $N$, let $\epsilon >0$ satisfy
$$ \epsilon \leq N^{-\{(\nu^{-1}+d^{-1})\left(\frac{d}{p} -s \right)_{+} + \frac{s}{d} \} } 
( \log N )^{-\frac{1}{\alpha} \left(\frac{d}{p} -s \right)_{+} -1 -\frac{s-\delta}{\alpha}} (\log (\log N) )^{\frac{s-\delta}{\alpha} } .$$ Moreover, let $W_1 \coloneqq 6dm(m+2)+4d+2$, 
$L = 4+3\lceil \log_2 (\frac{3^{d+1\vee m}}{ \epsilon c_{(d, m)} }) +5 \rceil \lceil \log_2 (d+1\vee m) \rceil, W=NW_1, S=[(L-1)W_1 ^2 +1]N$, \\
$A_N \coloneqq
N^{r \{ \frac{s}{d} + (\nu^{-1} +d^{-1}) \left(\frac{d}{p} -s \right)_{+} \} } (\log N)^{\frac{r}{\alpha}\left(\frac{d}{p} -s \right)_{+} + r} \left( \frac{ \log N }{ \log ( \log N ) } \right)^{\frac{1}{\alpha} ( -d+1+ sr - r\delta ) }  $, \\
$B_N\coloneqq N^{(\nu^{-1} +d^{-1}) \left(1 \vee \left( \frac{d}{p} -s \right)_{+} \right) } ( \log N )^{\frac{1}{\alpha} \left(1 \vee \left( \frac{d}{p} -s \right)_{+} \right)}$ and 
$B = O(A_N \vee B_N)$. It then holds that
\[ \sup_{f^\circ \in U_{B_{p, q}^{s(x)}(\Omega) }} \inf_{\hat{f} \in \Phi(L, W, S, B)} \| f^\circ - \hat{f} \|_r \lesssim N^{-\frac{s}{d}} \left( \frac{ \log N }{\log(\log N)} \right)^{-\frac{s-\delta }{\alpha }}. \]

\end{theorem}

\subsection{Estimation error of deep learning}\label{deep-estimation}

First, for the estimation error, we confirm that the lower bound of the polynomial factor is $n^{-\frac{2s_{\min}}{2s_{\min}+d}}$ if $X$ takes a value around the minimum point of $s(x)$ with a certain probability. Let $a$ satisfy $s(a) = s_{\min}$. We suppose that there exists a constant $t > 0$ such that $p(x)$ satisfies $\inf_{x \in Q} p(x) > 0$, where $Q\coloneqq [a-\frac{t}{2}, a-\frac{t}{2}]^d$. This assumption ensures that $X$ takes values in the domain where the estimation is most difficult with a certain probability. Under this assumption, we can obtain the inequality below:
\begin{align*} \expect \left[ \int_{\Omega} ( f(x) - \hat{f}(x) )^2 p(x) \mathrm{d}x \right] &\geq \expect \left[ \int_{Q} ( f(x) - \hat{f}(x) )^2 p(x) \mathrm{d}x \right] \\
&\asymp \expect \left[ \int_{Q} ( f(x) - \hat{f}(x) )^2 \mathrm{d}x \right], 
\end{align*}
where $\hat{f} : [0, 1]^d \rightarrow \zissu$ is the function that is estimated from the observed data $(X_i, Y_i)_{i=1}^{n}$ and the expectation is taken with respect to the observed data $(X_i, Y_i)_{i=1}^{n}$. Moreover, note that the following holds:
\[ \sup_{f \in U_{B_{p, q}^{s}(Q)} } \expect \left[ \int_{Q} ( f(x) - \hat{f}(x) )^2 \mathrm{d}x \right] \asymp \sup_{g \in U_{B_{p, q}^{s}(\Omega)} } \expect \left[ \int_{\Omega} ( g(x) - \hat{g}(x) )^2 \mathrm{d}x \right]. \]
Here, the transformation from $f$ into $g$ and from $\hat{f}$ into $\hat{g}$ is based on the translation and scale change of $x$. By the definition of a Besov space, it holds that $\| f \|_{B_{p, q}^{s}(Q)} \asymp \| g \|_{B_{p, q}^{s}(\Omega)}$. It is known that $\inf_{\hat{f} }R(\hat{f}, B_{p, q}^{s}(\Omega) ) \gtrsim n^{-\frac{2s}{2s+d}}$ \citep{kerkyacharian1992density, donoho1996density, donoho1998minimax, gine2016mathematical}; thus, by the same argument as \autoref{lower-bound}, for all $s(x)$ it holds that 
\[ ^\forall \epsilon > 0 , \quad \inf_{\hat{f} }R(\hat{f}, B_{p, q}^{s(x)}(\Omega) ) \gtrsim n^{-\frac{2(s_{\min}+\epsilon)}{2(s_{\min}+\epsilon)+d}}. \]
 Therefore, it is important to consider the poly-log order when determining the difference in the convergence rate.
 
Now, we evaluate only the $L_2$ norm risk, and thus introduce $\nu \coloneqq d\left( \frac{1}{p} - \frac{1}{2} \right)_{+}$ instead of $\delta$. 

To obtain the upper bound of the estimation error through a deep neural network, we use the following lemma.
\begin{lemma}[\citep{schmidt2020nonparametric, suzuki2018adaptivity}]\label{network} {The covering number of $\Phi(L, W, S, B)$ is bounded by
\begin{align}
 \log\mathcal{N}(\delta, \Phi(L, W, S, B) , \| \cdot \|_{\infty} ) &\leq S\log( \delta^{-1}L(B\vee 1)^{L-1} (W+1)^{2L} ) \notag \\ &\leq 2SL\log( (B\vee1)(W+1))+S\log(\delta^{-1}L). \notag
\end{align}
 }
 \end{lemma}
Before the proof of \autoref{estimation-error}, we introduce some notations. Let $F > 0$, and we define $\Psi(L, W, S, B)$ as follows:
\[ \Psi(L, W, S, B)\coloneqq \{\max\{ \min\{ f, F \} , -F \} \mid f \in \Phi( L, W, S, B )\}. \]
We can see that $ \Psi(L, W, S, B)$ is the clipping of $\Phi(L, W, S, B)$ by $F$. The clipping is easily realized by the ReLU function as follows: $\eta(x + F) - \eta(x - F) - F$.

\begin{theorem}\label{estimation-error}{Suppose $0<p, q \leq \infty$, $0 < s <\infty$ and $s > \nu$. If $f^{\circ} \in U_{ B_{p, q}^{s+\beta \| x - c \|_2 ^{\alpha} }(\Omega)}$ and $\| f^{\circ} \|_{\infty} \leq F$, where $F \geq 1$, letting $(L, S, W, B)$ be as in \autoref{approxi} with $N \asymp n^{\frac{d}{2s+d}} (\log n)^{-\frac{d(3\alpha+2s-2\nu)}{(2s+d)\alpha}} (\log(\log n))^{\frac{2d(s-\nu)}{(2s+d)\alpha} }$, it holds that
\[ \expect[\norumu{f^\circ(X) - \hat{f}(X) } {{L_2(P_X)}}^2] \lesssim n^{-\frac{2s}{2s+d}}(\log n)^{-\frac{2(sd-\nu d-3\alpha s)}{(2s+d)\alpha}}( \log(\log n) )^{\frac{2d(s-\nu)}{(2s+d)\alpha}}, \] }
where $\hat{f} \in \Psi(L, W, S, B)$ is the least squares estimator for $f^\circ $ whose definition is given in Eq.(\ref{least-square}). 
\begin{Proof}
By \autoref{approxi}, each parameter of the neural network satisfies $L = O( \log N ), W = O(N), S = O(N \log N)$, and $B = O(N^k (\log N )^l)$, where $k, l \in \zissu$. By applying \autoref{network}, it holds that
\[ \log\mathcal{N}(\delta, \Psi(L, W, S, B) , \| \cdot \|_{\infty} ) \lesssim N\log N\{ (\log N)^2 +  \log(\delta^{-1} )\}.\]
Here, by $\| f^\circ \|_{\infty} \leq F$, it holds that $\| f^\circ - \max\{ \min\{ \tilde{f}, F \}, -F\} \|_2 \leq \| f^\circ -\tilde{f} \|_2$. Therefore, by \autoref{approxi}, 
\[ \inf_{\tilde{f}\in \Psi(L, W, S, B) } \| f^{\circ}-\tilde{f} \|_{2} \lesssim N^{-\frac{s}{d}} \left( \frac{ \log N }{\log(\log N)} \right)^{-\frac{s-\nu }{\alpha }}.\]
In addition, because the density function $p(x)$ satisfies $p(x)\leq T$, 
\[ \inf_{\tilde{f}\in \Psi(L, W, S, B) } \| f^{\circ}-\tilde{f} \|_{L_2(P_X)} \lesssim N^{-\frac{s}{d}} \left( \frac{ \log N }{\log(\log N)} \right)^{-\frac{s-\nu }{\alpha }}. \]
By applying \autoref{shmid} with $\delta = \frac{1}{n}$, we have 
\[ \expect[ \| \hat{f} - f^{\circ} \|_{L^2(P_{X})} ^2] \lesssim N^{-\frac{2s}{d}} \left( \frac{ \log N }{\log(\log N)} \right)^{-\frac{2(s-\nu) }{\alpha }} +\frac{ N\log N\{ (\log N)^2 + \log n) \} }{n}+ \frac{1}{n}. \]
Here, we let $N$ satisfy 
$N \asymp n^{\frac{d}{2s+d}} (\log n)^{-\frac{d(3\alpha+2s-2\nu)}{(2s+d)\alpha}} (\log(\log n))^{\frac{2d(s-\nu)}{(2s+d)\alpha} }$; consequently, we can obtain the desired result. \\
\qed
\end{Proof}
\end{theorem}
For $p > 2$, the poly-log order of estimation error is $(\log n)^{-\frac{2s(d-3\alpha )}{(2s+d)\alpha}}$, and if $s=1$, it is $(\log n)^{-\frac{2(d-3\alpha )}{(2+d)\alpha}}$. Thus, we can see that the influence of the poly-log order increases as dimension $d$ increases, and as $\alpha$ decreases. By contrast, because the polynomial order is affected by the curse of dimensionality, we can also see that the influence of the poly-log order increases as the dimension increases.

\subsection{Numerical evaluation of the improvement of the estimation error by adaptive approximation}\label{computation}

\begin{figure}[H]
 \begin{minipage}{0.50\hsize}
  \centering
   \includegraphics[width=65mm]{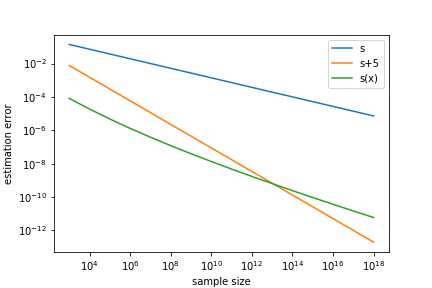}
  \subcaption{$\alpha=0.2, d=5$}
  \label{fig:one}
 \end{minipage}
 \begin{minipage}{0.50\hsize}
 \centering
  \includegraphics[width=65mm]{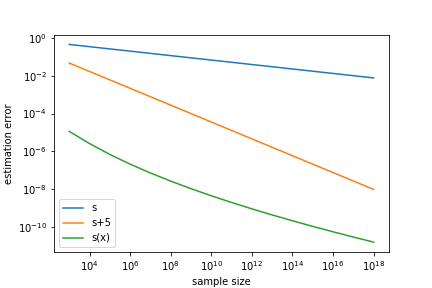}
  \subcaption{$\alpha=0.2, d=15$}
  \label{fig:two}
 \end{minipage}
 \begin{minipage}{0.50\hsize}
\centering
  \includegraphics[width=65mm]{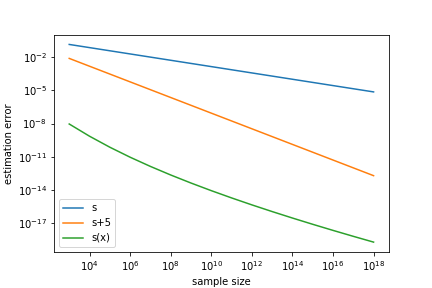}
  \subcaption{$\alpha=0.1, d=5$}
  \label{fig:three}
 \end{minipage}
  \caption{Numerical evaluation of the estimation error}\label{numerical}
\end{figure}

The estimation error is improved by the adaptive method, and we numerically evaluate the effect of the improvement. This numerical evaluation compares the estimation error for the realistic number of observations when the parameters change. We confirm that the improvement is significant if the parameters satisfy certain conditions. In particular, we will see that the poly-log factor is significant when $\alpha$ is small or $d$ is large. \autoref{numerical} represents the estimation error of deep learning on $B_{p, q}^{s+\beta \| x - c \|_2 ^\alpha}(\Omega)$ $\left( n^{-\frac{2s}{2s+d}}(\log n)^{-\frac{2(sd-\nu d-3\alpha s)}{(2s+d)\alpha}}( \log(\log n) )^{\frac{2d(s-\nu)}{(2s+d)\alpha}} \right)$ as green, the polynomial order of the estimation error of deep learning on $B_{p, q}^{s+5}(\Omega)$ $\left( n^{-\frac{2(s+5)}{2(s+5)+d}} \right)$ as orange, and that of $B_{p, q}^{s}(\Omega)$ $\left(n^{-\frac{2s}{2s+d}} \right)$ as blue on the log-log graphs with $s = 1$, $2\leq p$. In each graph, the horizontal line represents the number of observations, and the vertical line represents the estimation error. \autoref{fig:two} shows the case in which $d$ is larger than that of \autoref{fig:one}, and \autoref{fig:three} shows the case in which $\alpha$ is smaller that of \autoref{fig:one}.

 Note that we only need to consider the inclination of the graphs here because the constant factor is different in each graph. We can see that the improvement due to the adaptive method is significant if $d$ is large or $\alpha$ is small. Under this condition, we can consider that the order of the convergence rate for the estimation in $B_{p, q}^{s+\beta \| x - c \|_2 ^\alpha}(\Omega)$ is equivalent to that of $B_{p, q}^{s+5}(\Omega)$, if the number of observations is realistic. Therefore, we can see that if $\alpha$ is small or $d$ is large, the estimation error is far better than that of $B_{p, q}^{s}(\Omega)$.

\section{Superiority to linear estimator}\label{Superiority to linear estimator}

In this section, we show the superiority of deep neural networks to linear estimators by focusing on the adaptivity of such networks. A linear estimator is a class of estimators that is linearly dependent on outputs $(Y_1, Y_2, \ldots, Y_n)$. This class includes some popular methods in the field of machine learning, for example, linear regression, Nadaraya-Watson estimator, and kernel ridge regression. Because kernel methods can be considered as a learning method using shallow neural networks, we may regard this comparison as that between deep neural networks and shallow neural networks.

Here, we define a linear estimator as follows:
\begin{definition}
The linear estimator is a class of estimators that can be written as follows:
\[ \hat{ f }(x) = \sum_{i=1}^{n} Y_i \varphi_i (x, X^n), \]
where $X^n = (X_1, X_2, \ldots, X_n)$ and $\varphi_i (x, X^n)\ (i=1, \ldots, n)$ are measurable functions. 
\end{definition} 
This class includes the least squares estimator, Nadaraya-Watson estimator, and kernel ridge regression. For example, the estimator from kernel ridge regression can be written as follows:
\[ \hat{f}(x) = \bm{k}(x)^\top (K+\lambda I_n)^{-1} Y,\]
where $\lambda > 0$, and $Y = (Y_1, Y_2, \ldots, Y_n)^\top$, $k : \zissu^d \times \zissu^d \to \zissu$ is a positive semi-definite kernel, $\bm{k}(x) \coloneqq ( k(x, X_1), k(x, X_2), \ldots k(x, X_n) )^\top$, $K=(k(X_i, X_j))_{i, j} $. 
 Linear estimators were compared with deep learning in previous studies. It was shown that deep learning is superior to any linear estimator for specific target function classes $\mathcal{F}$ \citep{suzuki2018adaptivity, suzuki2019deep, hayakawa2020minimax, pmlr-v89-imaizumi19a}.
 
 Here, we introduce some properties of linear estimators. Whether the minimax optimal rate of a linear estimator is equivalent to that of a function space that is larger than the original target function class $\mathcal{F}$ was considered in previous studies. Indeed, it was proved in \cite{hayakawa2020minimax} that the optimal minimax rate of a linear estimator does not differ from that of a convex hull of the original target function class: 
 \begin{equation}\label{hayakawa-convex}
 \inf_{\hat{f}: \rm{linear}} \sup_{ f^\circ \in \mathcal{F}^\circ} E[ \| f^\circ - \hat{f} \|_{L_2 (P_X)} ^2 ] = \inf_{\hat{f}: \rm{linear}} \sup_{ f^\circ \in \rm{conv}(\mathcal{F}^\circ) } E[ \| f^\circ - \hat{f} \|_{L_2 (P_X)} ^2 ], 
 \end{equation}
where the infimum is taken over the linear estimators and $ \rm{conv}(\mathcal{F}^\circ)$ is defined as follows:
\[ \mathrm{conv}(\mathcal{F}^\circ) \coloneqq \left \{ \sum_{i=1}^{k} t_i f_i \ \middle| \ t_1, t_2, \ldots , t_k \geq 0, \sum_{i=1}^{k} t_i =1, f_1, f_2, \ldots, f_k \in \mathcal{F}^\circ , k\geq 1 \right \}.\] 
In addition, under certain assumptions, this was shown not only for the convex hull, but also for Q-hull \citep{donoho1990minimax, donoho1998minimax}.

We define a function set $\mathcal{G}$ as follows: 
\[ \mathcal{G} \coloneqq \bigcup_{c\in[0, 1]^d} U_{B_{p, q}^{s+\beta\| x-c\|_{2} ^\alpha}(\Omega)}. \]
Because we do not know the location of $c$ in estimating a function in $\mathcal{G}$, it is difficult to identify which part of the function is less smooth (hard to estimate). This setting is more natural than that in which the target is in $ U_{B_{p, q}^{s+\beta\| x-c\|_{2} ^\alpha}(\Omega)}$. We can see that the estimation in $\mathcal{G}$ requires more adaptivity to achieve better accuracy. In addition, if $p(x)$ satisfies the following condition,
\[\ there\ exists \ a \in \Omega \ and\ t > 0 \ such\ that\ \inf_{x \in [ a - t, a+ t ]^d} p(x) > 0, \]
for any estimator, the lower bound of a polynomial factor of the estimation error on $\mathcal{G}$ is $n^{-\frac{2s}{2s +d}}$.
 Thus, unless $P^{X}$ is a discrete measure, the lower bound of the polynomial factor of the estimation error is $n^{-\frac{2s}{2s+d} }$. 

The main theorem in this section is \autoref{linear-superior}. This indicates the superiority of a deep neural network over a linear estimator. The proof is provided in Appendix \ref{proof of linear superoor}.

\begin{theorem}\label{linear-superior}{Suppose that $0 < p \leq 2$, $0 < q \leq \infty$ and  $P_X$ has a uniform distribution. Then, if $s > \nu$, 
\[ \inf_{\hat{f}: \mathrm{linear}} \sup_{ f\in \mathcal{G} } \expect[ \| f -\hat{f} \|_{L_{2}(P_X)} ^2 ] \gtrsim n^{-\frac{2(s-\nu)}{2(s-\nu)+d} }. \] }

\end{theorem}
Note that, because the upper bound in \autoref{estimation-error} does not depend on the location of $c$, the upper bound of the estimation error of deep learning on $\mathcal{G}$ is $n^{-\frac{2s}{2s+d}}(\log n)^{-\frac{2(sd-\nu d-3\alpha s)}{(2s+d)\alpha}}( \log(\log n) )^{\frac{2d(s-\nu)}{(2s+d)\alpha}}$. Therefore, deep learning is superior to the linear estimators for $0 < p < 2$ or $p = 2$ and $0< \alpha < \frac{d}{3}$:
\[ n^{-\frac{2s}{2s+d}}(\log n)^{-\frac{2(sd-\nu d-3\alpha s)}{(2s+d)\alpha}}( \log(\log n) )^{\frac{2d(s-\nu)}{(2s+d)\alpha}}\ (\mathrm{deep}) \ll n^{-\frac{2(s-\nu)}{2(s-\nu)+d} }\ (\mathrm{linear}). \] 
This difference is due to the adaptivity. Although linear estimators cannot estimate adaptively the location of $c$ because the basis functions are fixed, deep learning can estimate the location of $c$ adaptively. In addition, it can be seen that the minimax optimal rate for the linear estimators do not have any additional poly-log order factor. That is, a linear estimator cannot induce a poly-log order improvement that was attained by the adaptive method. These results characterize the adaptivity of a deep neural network.

\begin{remark}
Let $\mathcal{F}$ be the target function class of the estimation. \cite{hayakawa2020minimax} showed that the minimax rate of a linear estimator is equivalent to that of a convex hull of $\mathcal{F}$. In addition, \citep{donoho1990minimax, donoho1998minimax} showed that under some proper conditions, it is equal to that of QHull of $\mathcal{F}$. Here, to provide an intuitive understanding of \autoref{linear-superior}, we present another proof of \autoref{linear-superior}.
First, we define the following set: 
\[ \mathcal{M} \coloneqq \left\{ t_{k, j} M_{k, j} ^d \ \middle| \ 0\leq | t_{k, j} | \leq 2^{-k(s-\frac{d}{p})} \ ( j\in \Lambda(k), k=0, 1, \ldots )\right\}. \]
It follows, by \autoref{bsp-hyouka}, that $\mathcal{M} \subset C_1 \mathcal{G}$ for some $C_1 >0$. By (\ref{norm-tenkai}), we have
\[ C_2 U_{ B_{1,1}^{s+d-\frac{d}{p}}(\Omega)} \subset \overline{  \mathrm{conv}( \mathcal{M} )}  ,  \]
for some $C_2 >0$. Note that the closure is taken with respect to the $L_2$ norm. By (\ref{hayakawa-convex}), the minimax optimal rate of the linear estimators for $\mathcal{G}$ does not exceed the minimax rate of the linear estimators of $B_{1,1}^{s+d-\frac{d}{p}}(\Omega)$. By proving it in the same way as \autoref{linear-superior}, it holds that the minimax rate of the linear estimators for $B_{1, 1}^{s+d-\frac{d}{p}}(\Omega)$  has a lower bound
\[ n^{-\frac{2(s+d -\frac{d}{p}-\frac{d}{2}) }{ 2(s+d -\frac{d}{p}-\frac{d}{2})+d }}= n^{-\frac{2(s-\nu)}{2(s-\nu)+d}}. \]
Therefore, \autoref{linear-superior} can be shown.

From the proof above, we can see that for the linear estimators, it is more difficult to estimate the functions in $\mathcal{G}$ than $B_{1,1}^{s+d-\frac{d}{p}}(\Omega)$. Because $B_{1,1}^{s+d-\frac{d}{p}}(\Omega)$ does not have any information about $s(x)$, linear estimators cannot be adaptive to the shape of $s(x)$. For $p > 2$, if $p$ is close to $2$ and $\alpha$ is small, by the argument of $\autoref{computation}$, it is expected that deep learning is superior to linear estimators for a realistic sample size. Of course, the convergence rate of the estimation error of a linear estimator on $B_{1, 1}^{s+\frac{d}{2}-\frac{d}{p} }(\Omega)$ is faster in a Landau symbol than that on $B_{p, q}^{s+\beta\|x - c \|_2 ^\alpha}(\Omega)$. 
\end{remark}

\section*{Related work}
We note that there are some papers that investigated  estimation problems for a function class with variable smoothness. In the context of kernel density estimation, \cite{gach2013spatially} and \cite{patschkowski2019locally} treated special H\"{o}lder spaces whose smoothness is locally defined. Although the measurement of the estimation error and treated function space differ from this paper, estimators in \cite{gach2013spatially} and \cite{patschkowski2019locally} adapt to the variable smoothness. The main idea that is common in both papers is adjusting bandwidth locally so as to adapt to the local smoothness. This idea is similar to that of this work which adjusts the resolution level of B-spline basis. However, there are many differences between those works and this work. \cite{gach2013spatially} and \cite{patschkowski2019locally}  only treated the case $d=1$. \cite{gach2013spatially}  restricted  $s(\cdot)$ to $0 < s(\cdot) \leq 1$ and Proposition 3.13 in \cite{patschkowski2019locally} which is consistent with \autoref{lower-bound} only considered the case  $0 < \inf_{x\in \Omega} s(x)  \leq 1$. In addition, we suggested the relation between  the poly-log improvement  of $L_2$ estimation error and $s(\cdot)$ and $d$ while Theorem 3.15 in \cite{patschkowski2019locally}
 gave the upper bound of estimation error in abstract form. Although those works proved the pointwise convergence which is stronger than $L_2$ convergence of this work, we proved  the convergence  for a broader function class which includes the function class equipped with locally H\"older smoothness. In addition, we gave the approximation  theory  of B-spline basis for a variable exponent Besov space which is broader than  locally H\"older space and we proved the superiority to linear estimators. 

The nonparametric regression problem on a Besov space $B_{p, q}^s (\Omega)$ was studied extensively and you may refer to Chapter 9, 10, 11 in \cite{hardle2012wavelets} and Chapter 4.3, Chapter 6.3 and Chapter 8.2 in \cite{gine2016mathematical}. In particular, adaptivity of wavelet threshold estimators has been known \citep{donoho1996density, donoho1998minimax}. Wavelet threshold estimators adapt to the spatially homogeneity of smoothness, and the nonlinearity of those estimators improve the convergence rate when the parameter $p$ is small. However, as far as we know, the performance of wavelet shrinkage estimators in the variable exponent setting has not been analyzed. We would like to defer this to a future work.
\section*{Conclusion}

We showed that the polynomial order of the approximation and estimation errors cannot be improved from the order of the minimum value of $s(x)$ and that the adaptivity of deep learning yields a poly-log order improvement. This improvement is remarkable when the dimension is large and the area around the minimum value of $s(x)$ is small, that is, the domain, where the estimation is the most difficult, is small. In addition, we have shown that, for $0 < p \leq 2$, no linear estimator can achieve the poly-log improvement, and we can ensure the superiority to linear estimators with respect to the estimation error. Notably, these results provide insight into the high performance of deep learning in the application fields. 

\section*{Acknowledgment}
I would like to thank Sho Sonoda, Koichi Taniguchi, Masahiro Ikeda, Mitsuo Izuki, and Takahiro Noi for the discussions. TS was partially supported by JSPS KAKENHI (18K19793, 18H03201, and 20H00576), Japan DigitalDesign, and JST CREST. We would like to thank Editage (www.editage.com) for English language editing.

\bibliographystyle{apalike}        
\bibliography{thesis_submitted.bib}

\appendix
\section{Proofs}
\subsection{Proof of \autoref{sup-th2}}\label{proof-of-sup}
\begin{Proof}
(i) Suppose $p \geq r$. 

 It holds that $s(x) > s+\beta t^{\alpha}$, where $x\in A^c$. Thus, by (\ref{Dung's inequality}) we have the following: 
\[ \| f \mathbbm{1}_{A^c} - Q_{\bar{k} }\mathbbm{1}_{A^c} \|_r \lesssim 2^{-\bar{k}(s+\beta t^{\alpha} ) }. \]
Moreover, by (\ref{Dung's inequality}), it holds that
\[ \| f\mathbbm{1 }_A - Q_{\bar{k}+N_k} \mathbbm{1 }_A\|_r \lesssim 2^{-(\bar{k}+N_k)s}. \]
Therefore, 
\[ \| f -f_N \|_r \lesssim \| f \mathbbm{1}_{A^c} - Q_{\bar{k} }\mathbbm{1}_{A^c} \|_r + \| f\mathbbm{1 }_A - Q_{\bar{k}+N_k}\mathbbm{1 }_A \|_r \lesssim 2^{-\bar{k}(s+\beta t^{\alpha} ) } +2^{-(\bar{k}+N_k)s}. \]

(ii) Suppose $ p < r $.

Because it holds that $s(x) > s+\beta t^{\alpha}$ with $x\in A^c$, by applying the proof of Theorem 3.1 in \cite{dung2011optimal}, we have
\begin{equation}\label{thm2-2-1}
 \| f \mathbbm{1}_{A^c} - Q_{\bar{k} }\mathbbm{1}_{A^c} - \sum_{k=\bar{k}+1}^{k^*} \sum_{j=1}^{n_k} a_{k, v_{A^c, j} } M_{k, v_{A^c, j} }^{d} \mathbbm{1}_{A^c} \|_r \lesssim 2^{-\bar{k}(s+\beta t^{\alpha} ) }. 
\end{equation}

Next, we consider the case of $x \in A$.
By (\ref{tenkai}), $f \in B_{p, q}^{s(x)}$ can be decomposed as follows:
\[ f (x) = \sum_{k=0}^{\infty} q_k (f). \]
Note that, by the embedding $B_{p, q}^{s}(\Omega) \hookrightarrow B_{r, q}^{s-\delta}(\Omega)$, if $s > \delta$, it converges with respect to the $L_r$ norm. We define $q_{k, A} (f)$ as $q_{k, A} (f) \coloneqq \sum_{j \in \Lambda_{A} (k) } a_{k, j}M_{k, j} ^d$. We also define sequence $\{ a^{*}_{k, v_{A, j}} \}$ as follows:
\[
  a^{*}_{k, v_{A, j}} = \begin{cases}
    a_{k, v_{A, j}} & (1\leq v_{A, j} \leq m_k), \\
   0 & (m_k \leq v_{A, j}).
  \end{cases}
\]
By (\ref{norm-tenkai}), the following holds:
\[ \| q_{k, A} - G_k(q_k , m_k, A)\|_r \asymp 2^{-\frac{d k}{r} } \| a_{k, v_{A, j} } - a^{*}_{k, v_{A, j}} \|_{\ell_r}. \]
Moreover, by Lemma 5-1 in \cite{dung2011optimal}, we have 
\[ \| a_{k, v_{A, j} } - a^{*}_{k, v_{A, j}} \|_{\ell_r} \lesssim m_k ^{-\frac{\delta}{d}} \| a_{k, v_{A, j} } \|_\ell{_p}.\]
Combining the two formulas above with $ 2^{-\frac{dk}{p} } \| a_{k,v_{A, j} } \|_{\ell_p} \lesssim 2^{-sk} $, it holds that 
\begin{equation}\label{thm2-2-2}
 \|q_{k, A} - G_k(q_k , m_k, A) \|_r \lesssim 2^{-sk} 2^{ \delta k} m_k ^{-\frac{\delta}{d}}. 
 \end{equation}
Moreover, for $0 < \tau \leq \min\{1, r \}$ and the function sequence $\{ f_k \}$, it holds that
\begin{equation}\label{thm2-2-3}
\| \sum f_k \|_r ^\tau \leq \sum \| f_k \|_r ^\tau.
\end{equation}
By applying (\ref{norm-tenkai}), embedding $B_{p, q}^{s}(\Omega) \hookrightarrow B_{r, q}^{s-\delta}(\Omega)$, (\ref{thm2-2-2}) and (\ref{thm2-2-3}), we have the following inequality: 
\begin{align}
 &\| f\mathbbm{1}_{A} - Q_{\bar{k}+N_k }(f) \mathbbm{1}_{A} - \sum_{k=\bar{k} +N_k +1}^{(\bar{k}+N_k)^{*}} \sum_{j=1}^{m_k} a_{k, v_{A, j}} M_{k, v_{A, j}}^{d} \mathbbm{1}_{A} \|_r ^{\tau} \notag\\
&\lesssim \sum_{k=\bar{k} + N_k +1}^{ (\bar{k}+N_k)^{*} } \| q_{k, A} (f) -G_k(q_k, m_k, A) \|_r ^\tau + \sum_{(\bar{k}+N_k)^{*} \leq k} \| q_k \|_r ^\tau \notag \\
&\lesssim \sum_{k=\bar{k} + N_k +1}^{ (\bar{k}+N_k)^{*} } 2^{-\tau s k}2^{\tau \delta k}m_k ^{-\tau \frac{\delta}{d} }+ \sum_{(\bar{k}+N_k)^{*} \leq k} 2^{-\tau sk}2^{\tau\delta k} \notag\\
&\lesssim 2^{-\bar{k} \tau \delta} 2^{-\tau(s-\delta)(\bar{k}+N_k) }\sum_{ k = \bar{k}+N_k +1 }^{ (\bar{k}+N_k)^{*} } 2^{-\tau(s - \delta - \epsilon \frac{\delta}{d})(k - \bar{k}- N_k )} \notag \\
&\quad + \sum_{(\bar{k}+N_k)^{*} \leq k} 2^{-\tau sk}2^{\tau\delta k} \notag \\
&\lesssim 2^{-\tau \delta \bar{k} } 2^{-\tau(s-\delta)(\bar{k} +N_k) }+2^{-\tau(s-\delta)(\bar{k}+N_k)^{*}} \notag \\
& \lesssim 2^{-\tau s \bar{k}} 2^{-\tau (s-\delta)N_k}. \notag
\end{align}
Thus, we obtain 
\begin{equation}\label{thm2-2-4}
 \| f\mathbbm{1}_{A} - Q_{\bar{k}+N_k }(f) \mathbbm{1}_{A} - \sum_{k=\bar{k} +N_k +1}^{(\bar{k}+N_k)^{*}} \sum_{j=1}^{m_k}  a_{k, v_{A, j}} M_{k, v_{A, j}}^{d} \mathbbm{1}_{A} \|_r \lesssim 
2^{- s \bar{k}} 2^{- (s-\delta)N_k}.
\end{equation}
 Therefore, by (\ref{thm2-2-1}) and (\ref{thm2-2-4}), it holds that \\
\begin{align}
 &\| f- f_N \|_r \notag\\
 &\lesssim \| f \mathbbm{1}_{A^c} - Q_{\bar{k}}\mathbbm{1}_{A^c} - \sum_{k=\bar{k}+1}^{k^*} \sum_{j=1}^{n_k} a_{k, v_{A^c, j}} M_{k, v_{A^c, j}} \mathbbm{1}_{A^c} \|_r 
\notag\\&+ \| f\mathbbm{1}_{A} - Q_{\bar{k}+N_k }(f) \mathbbm{1}_{A} - \sum_{k=\bar{k} +N_k +1}^{(\bar{k}+N_k)^{*}} \sum_{j=1}^{m_k} a_{k, v_{A, j}} M_{k, v_{A, j}} \mathbbm{1}_{A} \|_r \notag\\
&\lesssim 2^{-\bar{k}(s+\beta t^{\alpha} ) } + 2^{- s \bar{k}} 2^{- (s-\delta)N_k}. \notag
\end{align}

\qed
\end{Proof}
\begin{remark}
We used the following property in the proof of \autoref{sup-th2}: If domain $A$ is a minimally smooth domain and $\inf_{x\in A} s(x) > s$, $\| f-  Q_k \|_r \lesssim 2^{-k(s-\delta)}$. This is followed by applying \autoref{extension}, and it holds that for all $f \in U_{B_{p, q}^{s(x)}(\Omega) }$ there exists $ C >0 $ and $g \in C U_{B_{p, q}^{s}(\Omega) }$ that coincide with $f$ for all $x\in A$. Additionally, it holds that $\| g \|_{B_{p, q}^{s}(\Omega)} \leq C \| f \|_{B_{p, q}^{s}(A)}$.

\end{remark}

\subsection{Proof of \autoref{approxi}}\label{proof-approxi}
Before the proof of \autoref{approxi}, we prove the following lemma. 
 \begin{lemma}\label{tankin}{Let $\| f \|_{\infty}\leq F$ and $A= [c - t, c + t]^d $. We define $g_i (x_i)$ as 
 \begin{align*}
  g_i (x_i)= &\eta\left( \frac{1}{\xi} ( x_i - c_i +t)+1 \right) - \eta \left( \frac{1}{\xi} ( x_i - c_i +t) \right) - \eta \left( \frac{1}{\xi} ( x_i - c_i - t) \right) \\
  &+ \eta \left( \frac{1}{\xi} ( x_i - c_i - t) -1 \right) 
  \end{align*}
with $\xi \leq \min\left\{ \frac{\epsilon^r}{F^r t^{d-1}(d+1) }, \frac{t}{2^d}\right\}$. Let $g(x) = \prod_{i=1}^{d} g_i (x_i)$, and it then holds that
 $\| f g \|_{L_r(A^c)} \leq \epsilon$. 
 }
 \begin{Proof}
The Lebesgue measure of the area where $g(x) \neq 0$ with $x \in A^c$ is 
 \[ (t+\xi)^d - t^d = t^d \left\{ \left(1+\frac{\xi}{t} \right)^d -1 \right\}.\] 
 If $\xi \leq \frac{t}{2^d}$, we have $(1+ \frac{\xi}{t})^d \leq 1 + d\frac{\xi}{t} +2^d (\frac{\xi}{t})^2 \leq 1+(d+1)\frac{\xi}{t}$. Thus, it holds that 
 \[ (t+\xi)^d - t^d \leq t^d (d+1)\frac{\xi}{t} = t^{d-1} (d+1) \xi.\] 
Therefore,
 \[ \| f g \|_{L_r(A^c)} ^r \leq F^r t^{d-1} (d+1) \xi \leq \epsilon ^r. \]

 \qed
 \end{Proof}

 \end{lemma}
 
\begin{figure}[H]
\centering
\begin{tikzpicture}[scale= 4.0]
\node at (0,0) [below left] {$\mathrm{O}$};		
\draw [thick, ->](-0.05, 0)--(1.1,0) node [right]{$x_i$};		
\draw [thick, ->] (0,-0.05)--(0,1.2) node [above] {$y$};	
\draw [thick, domain=0.10:0.20] plot(\x,  {  10*(\x - 0.1)  }  ) ; 
\draw [thick, domain=0.20:0.60] plot(\x,  {1}  ) ;
\draw [thick, domain=0.60:0.70] plot(\x,  {  -10*(\x - 0.7)  }  ) ;
\draw [dashed](1,0) node[below] {$1$}--(1,1)--(0,1) ;		
\node at (0,1) [above left] {$1$};
\end{tikzpicture}
\caption{Graph of $g_i (x_i)$ }
\end{figure}
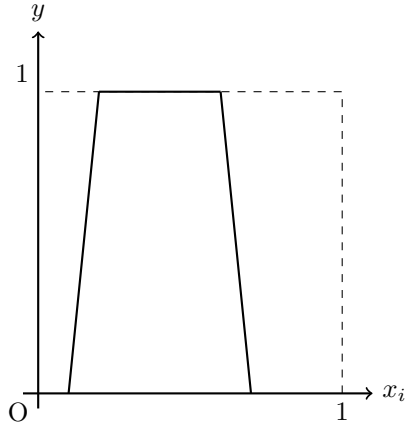

By using \autoref{tankin}, we prove \autoref{approxi}. 

\begin{Proof}
Note that $\| f \|_{B_{p, q}^{s}(\Omega)} \leq \| f \|_{B_{p, q}^{s(x)}(\Omega) }$. For $f\in U_{B_{p, q}^{s(x)}(\Omega)}$, by applying (\ref{norm-tenkai}), we have 
\begin{equation}\label{thm4-1-1}
|a_{k, j}| \lesssim 2^{k \left(\frac{d}{p} -s\right)_{+}}, 
\end{equation}
where $f (x) = \sum_{k=0}^{\infty} \sum_{j\in \Lambda(k)} a_{k, j} M_{k, j}^d (x)$. For $t$, which appears in the proof of \autoref{upper-bound}, we define $A$ as $A = [ c -t, c + t]^d$. We use the approximation of multiplication by the deep neural network below \citep*{yarotsky2017error, schmidt2020nonparametric, suzuki2018adaptivity}.\\
For all $\epsilon > 0$, let $L = \lceil \log_2 \left( \frac{3^D}{\epsilon} \right) +5 \rceil \lceil \log_2 (D) \rceil$, $W = 6d$, $S = LW^2$ and $B = 1$. Then, there exists $\phi_{\rm{mult}}(x_1, x_2, \ldots, x_D) \in \Phi(L, W, S, B)$ that satisfies the following two conditions:
\begin{align}
&1. \sup_{x \in [0, 1]^D} | \phi_{\rm{mult}}(x_1, x_2, \ldots, x_D) - \prod_{i=1}^D x_i | \leq \epsilon , \notag\\
&2. ^\exists i \in \{ 1, 2, \ldots, D \}, x_i = 0 \Rightarrow \phi_{\mathrm{mult}}(x_1, x_2, \ldots, x_D) = 0. \notag
\end{align}
By applying this to $\bar{M}$ in \autoref{B-spappro} and $g_i(x_i)= \eta( \frac{1}{\xi} ( x_i - c_i +t)+1 ) - \eta( \frac{1}{\xi} ( x_i - c_i +t) ) - \eta( \frac{1}{\xi} ( x_i - c_i - t) ) + \eta( \frac{1}{\xi} ( x_i - c_i - t) -1 ) $, we have
\[ \| \bar{M}g - \phi_{\mathrm{mult}} ( \bar{M}, g_1, g_2, \ldots , g_d ) \|_{\infty} \leq \epsilon, \]
where $g=\prod_{i=1}^{d}g_i , c(d, m) =  2+ 2de \frac{ (2 e )^m }{ \sqrt{m} } $, 
$L_1 \coloneqq 3+3\lceil \log_2 (\frac{3^{d+1\vee m}}{ \epsilon c_{(d, m)} }) +5 \rceil \lceil \log_2 (d+1\vee m) \rceil, W_1 \coloneqq 6dm(m+2)+4d+2, S_1\coloneqq L_1 {W_1}^2$ and $B_1 \coloneqq \max\{ 2(m+1)^m , \frac{1}{\xi} \}$. We define $\tilde{M}_A \in \Phi(L_1, W_1, S_1, B_1)$ as 
\[ \tilde{M}_A\coloneqq \phi_{\mathrm{mult}} ( \bar{M}, g_1, g_2, \ldots, g_d ) ,\]
 and $\tilde{M}_A$ satisfies the following condition: 
\begin{equation}\label{thm4-1-2}
\| M_{0,0} ^d g - \tilde{M}_A \|_{\infty} \leq \epsilon , \hspace{0.3cm} ^\exists i ,| x_i - c_i | \geq t+\xi \Rightarrow \tilde{M}_A (x)=0, 
\end{equation}
where $c = (c_1, c_2, \ldots, c_d)$.
We can construct $\tilde{M}_{A^c}$ in the same manner from ${g_i}^{'}(x_i) = 1- \eta( \frac{1}{\xi} ( x_i - c_i +t) ) + \eta( \frac{1}{\xi} ( x_i - c_i +t) - 1 ) + \eta( \frac{1}{\xi} ( x_i - c_i - t) +1 ) - \eta( \frac{1}{\xi} ( x_i - c_i - t) ) $.  $\tilde{M}_{A_{k, j}}$ and $\tilde{M}_{A^c_{k, ,j}}$ are defined in the same way as $M_{k, j}$. 

 Let $f_N$ be the function that appears in \autoref{upper-bound}, the set of indexes $(k, j)$ that consist of $f_N$ be $E_N$, and the set of indexes $(k, j)$ that consist of $Q_{\bar{k}+N_k }(f) \mathbbm{1}_{A}, \sum_{k=\bar{k} + N_k +1}^{(\bar{k} + N_k)^*} \sum_{j=1}^{m_k} a_{k, v_{A, j}} M_{k, v_{A, j}}^{d} \mathbbm{1}_{A} $ and satisfy $\mathrm{supp} M_{k, j}^{d} \cap \partial A \neq \emptyset $ be $E_{A_N}$. In the same manner, we define $E_{A^{c}_N}$. Further, we similarly define $E_{B} = E_N \backslash ( E_{A_N} \cup E_{A^{c}_N} )$. We define $\tilde{f}$ as follows:
\[ \tilde{f} = \sum_{(k, j)\in E_{A_N} } a_{k, j} \tilde{M}_{A_{k, j}} + \sum_{(k, j)\in E_{A^{c}_N} } a_{k, j} \tilde{M}_{{A^c}_{k, j}} + \sum_{(k, j)\in E_{B}} a_{k, j} \tilde{M}_{k, j}. \]
Note that for $(k, j)\in E_B$, we denote $\tilde{M}_{k, j}$ by 
\[\tilde{M}_{k, j}\coloneqq
\begin{cases}
\tilde{M}_{A_{k, j}} &(\mathrm{supp} M_{k, j} \subset A),\\
\tilde{M}_{A^c_{k, ,j}} & (\mathrm{supp} M_{k, j} \subset A^c).
\end{cases}
\]
In addition, for the simplicity of notation, we allow to denote $\tilde{M}_{A_{k, j}} $ and $\tilde{M}_{A^c _{k, j}}$ by $\tilde{M}_{k, j}$. we let 
\[ f_{1} =\mathbbm{1}_{A} \sum_{(k, j)\in E_{A_N} } a_{k, j} \tilde{M}_{A_{k, j}} + \mathbbm{1}_{A^c} \sum_{(k, j)\in E_{A^{c}_N} } a_{k, j} \tilde{M}_{{A^c}_{k, j}} + \sum_{(k, j)\in E_{B}} a_{k, j} \tilde{M}_{k, j}, \]
and evaluate the error owing to the approximating indicator functions. First, by applying (\ref{thm4-1-1}) and the properties of $M_{k, j}^d$, we have the following inequality:
\begin{align}
| f_{N} (x) - f_{1}(x) | &\leq \sum_{(k, j)\in E_N} | \alpha_{k, j} | | M_{k ,j}^{d} - \tilde{M}_{k, j} (x) | \notag \\
&\leq \epsilon \sum_{ (k, j)\in E_N } | \alpha_{k, j} | \mathbbm{1}\{ M_{k, j}^d (x) \neq 0 \} \notag\\
&\lesssim \epsilon (m+1)^d 2^{(\bar{k}+N_k)^{*} \left( \frac{d}{p} -s \right)_{+}} \{ 1+(\bar{k}+N_k)^{*} \} \| f \|_{B_{p, q}^{s} }, \label{thm4-1-5}
\end{align}
 where $\bar{k}+N_k = \bar{k} + \left \lceil \log \left( \frac{ \bar{k} }{ \log( \frac{\bar{k} }{ \log \bar{k} } )^{\frac{s-\delta}{\alpha} } } \right)^{\frac{1}{\alpha}} \right \rceil$. 
In addition, by definition, it holds that
 \[ (\bar{k}+N_k )^{*} = \left[ \nu^{-1} \left( \log\lambda +  \bar{k} d \right) \right]+ \bar{k} + \left \lceil \log\left( \frac{ \bar{k} }{ \log( \frac{\bar{k} }{ \log \bar{k} } )^{\frac{s-\delta}{\alpha} } } \right)^{\frac{1}{\alpha}} \right \rceil. \]
Through a simple calculation, it holds that $1\leq \log \left( \frac{\bar{k} }{ \log \bar{k} } \right)$. Thus, 
\[ \frac{s-\delta}{\alpha} \leq \log \left( \frac{\bar{k} }{ \log \bar{k} } \right)^{\frac{s-\delta}{\alpha} }.\]
By applying this inequality, for a constant $C \in \zissu$, we have
\begin{align}
(\bar{k}+N_k )^{*} 
&\leq \left[ \nu^{-1} \left( \log\lambda +  \bar{k} d \right) \right]+ \bar{k} + \left \lceil \log \left( \frac{ \bar{k} }{\frac{s-\delta}{\alpha}} \right) ^{\frac{1}{\alpha}} \right \rceil \notag\\
&= C+ (d\nu^{-1}+1 )\bar{k} + \frac{1}{\alpha}  \log \bar{k}. \label{thm4-1-3}
\end{align}
Therefore, (\ref{thm4-1-5}) is bounded as follows:
\begin{align}
(\ref{thm4-1-5})
&\lesssim 
\epsilon \times 2^{ ( \frac{d}{p} -s )_{+} \{ (d\nu^{-1}+1 )\bar{k} + \frac{1}{\alpha} \log \bar{k} \} } \{ (d\nu^{-1}+1 )\bar{k} + \frac{1}{\alpha} \log \bar{k} \} \notag\\
&\lesssim \epsilon \times N^{(\nu^{-1} +d^{-1}) \left(\frac{d}{p} -s \right)_{+}} ( \log N )^{\frac{1}{\alpha} \left(\frac{d}{p} -s \right)_{+} +1 }. \notag
\end{align}

By taking $\epsilon$ to satisfy 
\[ \epsilon\{ N^{(\nu^{-1} +d^{-1}) \left(\frac{d}{p} -s \right)_{+}} ( \log N )^{\frac{1}{\alpha} \left(\frac{d}{p} -s \right)_{+} +1 } \} \leq N^{-\frac{s}{d}} \left( \frac{ \log N }{\log(\log N)} \right)^{-\frac{s-\delta}{\alpha }}\]
 (i.e., 
$ \epsilon \leq N^{-\{(\nu^{-1}+d^{-1})\left(\frac{d}{p} -s \right)_{+} + \frac{s}{d} \} } 
( \log N )^{-\frac{1}{\alpha} \left(\frac{d}{p} -s \right)_{+} -1 -\frac{s-\delta}{\alpha}} (\log (\log N) )^{\frac{s-\delta}{\alpha} } $), it holds that
\begin{equation}\label{thm4-1-4}
 \| f_N - f_1 \|_{r} \lesssim N^{-\frac{s}{d}} \left( \frac{ \log N }{\log(\log N)} \right)^{-\frac{s-\delta }{\alpha }}. 
 \end{equation}\

Next, we evaluate the norm constant $B$. By applying (\ref{thm4-1-1}) and (\ref{thm4-1-3}), we have $$|a_{k, j} |\lesssim N^{(\nu^{-1} +d^{-1}) \left(\frac{d}{p} -s \right)_{+}} ( \log N )^{\frac{1}{\alpha} \left(\frac{d}{p} -s \right)_{+}}.$$ Comparing this with $2^{(k+N_k)^{*}}$, we have
\[ B \lesssim N^{(\nu^{-1} +d^{-1}) \left(1 \vee \left( \frac{d}{p} -s \right)_{+} \right) } ( \log N )^{\frac{1}{\alpha} \left(1 \vee \left( \frac{d}{p} -s \right)_{+} \right)}. \]

Next, we evaluate the error by approximating the indicator functions, that is, $\| \tilde{f} - f_1 \|_{r}$. First, we consider $E_{A_N}$. By a triangle inequality, 
\begin{align*}
 \| \sum_{ (k, j)\in E_{A_N} } a_{k, j} \tilde{M}_{A_{k, j}}\|_{L_r (A^c)} 
\lesssim &\| \sum_{ (k, j)\in E_{N_A} } a_{k, j} (\tilde{M}_{A_{k, j}}- M_{k, j}^d g) \|_{L_r (A^c)} \\
&+\| \sum_{ (k, j)\in E_{A_N} } a_{k, j} M_{k, j} ^d g \|_{L_r (A^c)}.
\end{align*}
Because $ \|  \tilde{M}_{A_{k, j}}  - M_{k, j}^d  g \|_{\infty} \leq \epsilon $, the first term is bounded by 
\[ \| \sum_{ (k, j)\in E_{A_N} } a_{k, j} (\tilde{M}_{A_{k, j}} - M_{k, j} ^d  g) \|_{L_r (A^c)} \leq \epsilon \sum_{ (k, j)\in E_N } | a_{k, j} | \mathbbm{1}\{ M_{k, j}^d (x) \neq 0 \}. \]
 This term is already bounded in the process to obtain (\ref{thm4-1-4}). The second term is bounded with respect to $\| \cdot \|_{\infty}$ by being applied in the same way as (\ref{thm4-1-5}) and (\ref{thm4-1-3}): 
\begin{align}
\| \sum_{ (k, j)\in E_{A_N} } a_{k, j} M_{k, j}^d \|_{\infty} 
&\lesssim (m+1)^d 2^{(\bar{k}+N_k)^{*} \left( \frac{d}{p} -s \right)_{+}} \{ 1+(\bar{k}+N_k)^{*} \} \| f \|_{B_{p, q}^{s} } \notag\\
&\lesssim \{ N^{(\nu^{-1} +d^{-1}) \left(\frac{d}{p} -s \right)_{+}} ( \log N )^{\frac{1}{\alpha} \left(\frac{d}{p} -s \right)_{+} +1 }. \notag
\end{align}
By taking each variable in \autoref{tankin} to satisfy $\epsilon = N^{-\frac{s}{d}} \left( \frac{ \log N }{\log(\log N)} \right)^{-\frac{s-\delta }{\alpha }}$,\\ $F= N^{(\nu^{-1} +d^{-1}) \left(\frac{d}{p} -s \right)_{+}} ( \log N )^{\frac{1}{\alpha} \left(\frac{d}{p} -s \right)_{+} +1 } , t= \left( \frac{ \log (\log N ) }{\log N} \right)^{\frac{1}{\alpha}}$, and 
$$ \frac{1}{\xi} = O\left( N^{r \{ \frac{s}{d} + (\nu^{-1} +d^{-1}) \left(\frac{d}{p} -s \right)_{+} \} } (\log N)^{\frac{r}{\alpha}\left(\frac{d}{p} -s \right)_{+} + r} \left( \frac{ \log N }{ \log ( \log N ) } \right)^{\frac{1}{\alpha} ( -d+1+ sr - r\delta ) } \right), $$
 it holds that
\begin{equation}\label{thm4-1-7}
 \| \sum_{ (k, j)\in E_{A_N} } a_{k, j} M_{k, j}^d g \|_{L_r (A^c)} \lesssim N^{-\frac{s}{d}} \left( \frac{ \log N }{\log(\log N)} \right)^{-\frac{s-\delta }{\alpha } }. 
\end{equation}
Following the same argument, we can obtain the same evaluation for $E_{A^c _N}$. By \autoref{upper-bound}, (\ref{thm4-1-2}), and (\ref{thm4-1-7}), we have the following inequality:
\[ \| f- \tilde{f} \|_{r} \lesssim \|f-f_N\|_{r} + \| f_N- f_1\|_{r}+ \| \tilde{f} - f_1 \|_{r} \lesssim N^{-\frac{s}{d}} \left( \frac{ \log N }{\log(\log N)} \right)^{-\frac{s-\delta} {\alpha }}. \]
Depth $L$ is $L = L_1+1$ because the approximation function is realized by the linear combination of the approximation functions of B-spline. Moreover, it is clear that width $W$ is $O(NW_1)$ and sparsity $S$ is $(L-1)W_{1} ^2 N +N$ because the last layer combines all approximation functions of B-spline.
\qed
\end{Proof}

\subsection{Proof of \autoref{linear-superior} } \label{proof of linear superoor}
\begin{lemma}\label{bsp-hyouka}{$ \| M_{k, 0}^d \|_{B_{p, q} ^{s+\beta \| x \|_2 ^{\alpha}}(\Omega)} \lesssim 2^{k(s-\frac{d}{p} )}$.}

\begin{Proof}
For simplicity of notation, we denote $M_{k, 0}^d$ by $f$. We let $0 < u <1$ and suppose that $ \sqrt{d} (m+1) 2^{-k} \leq 2^{-ku}$. Note that $0 < q < 1$, and by a triangle inequality, 
\[ \left(\int_{0}^{1}{(\omega_{r, p}^{*}(f, t) )^{q}\bunsu{1}{t}} \mathrm{d}t \right)^{\frac{1}{q}} 
\lesssim \left(\int_{0}^{2^{-ku}}{(\omega_{r, p}^{*}(f, t))^{q}\bunsu{1}{t}} \mathrm{d}t \right)^{\frac{1}{q}} +\left(\int_{2^{-ku}}^{1}{(\omega_{r, p}^{*}(f, t) )^{q}\bunsu{1}{t}} \mathrm{d}t \right)^{\frac{1}{q}}.
\]
First, we evaluate the first term. For $\|h\|_2 \leq 2^{-ku}$, note that $\Delta_{h}^r (f)(x) = 0$, where $\| x\|_2 \geq 2^{-ku}(r+1)$, it holds that 
\[ \omega_{r, p}^{*}(f, t) \leq t^{-\{s+ \beta(2^{-ku}(r+1) )^{\alpha} \} } \sup_{h\in \zissu :\norumu{h}{2}\leq t} \| \Delta_{h}^r (f) \|_p , \]
where $t \leq 2^{-ku}$. By the definition of a Besov space, 
\begin{equation}\label{lem5-1-1}
 \left(\int_{0}^{2^{-ku}}{(\omega_{r, p}^{*}(f, t) )^{q}\bunsu{1}{t}} \mathrm{d}t \right)^{\frac{1}{q}} \leq | f |_{ B_{p, q}^{s+ \beta(2^{-ku}(r+1) )^{\alpha} }(\Omega)}. \end{equation}
Next, we evaluate the second term. For $t \geq 2^{-ku}$, by $ \sqrt{d} (m+1) 2^{-k} \leq 2^{-ku}$, we have $\| \Delta_{h}^{r}(f) \|_{\infty} \leq \| f \|_{\infty}\leq 1$ where $2^{-ku} \leq \|h\|_2 \leq 1$. Thus, 
\[ \| t^{-s(x)}\Delta_{h}^{r}(f) \|_{p} \leq t^{-s_{\max}} 2^{-k \frac{d}{p} } (m+1)^\frac{d}{p}. \] 
Therefore, we have the following inequality:
\[ \omega_{r, p}^{*}(f, t) \leq {\max\{ \omega_{r, p}^{*}(f, 2^{-ku}), t^{-s_{\max}} 2^{-k \frac{d}{p} } (m+1)^\frac{d}{p} }\}.
\]
By the definition of a Besov space, 
\begin{equation}\label{lem5-1-2}
 \left(\int_{2^{-ku}}^{1}{(\omega_{r, p}^{*}(f, 2^{-ku}) )^{q}\bunsu{1}{t}} \mathrm{d}t \right)^{\frac{1}{q}} \leq | f |_{ B_{p, q}^{s+ \beta(2^{-ku}(r+1) )^{\alpha} }(\Omega)}. 
 \end{equation}
Here, we retake $u$ to satisfy $u < \frac{s}{s_{\max}}$, and it holds that
\begin{equation}\label{lem5-1-3}
 \left(\int_{2^{-ku}}^{1}{( t^{-s_{\max}} 2^{-k \frac{d}{p} } (m+1)^\frac{d}{p} )^{q}\bunsu{1}{t}} \mathrm{d}t \right)^{\frac{1}{q}} \leq C 2^{kus_{\max}}2^{-k\frac{d}{p}} < C 2^{k(s-\frac{d}{p})}, 
 \end{equation}
where $C$ is a constant that does not depend on $k$. By (\ref{norm-tenkai}), 
\[ | f |_{ B_{p, q}^{s+ \beta(2^{-ku}(r+1) )^{\alpha} }(\Omega)} \lesssim 2^{k( {s+ \beta(2^{-ku}(r+1) )^{\alpha} } -\frac{d}{p}) }. \]
By $ \lim_{k \to \infty} 2^{k\beta(2^{-ku}(r+1) )^{\alpha}} = 1 $, for a sufficiently large $k$, 
\begin{equation}\label{lem5-1-4}
 | f |_{ B_{p, q}^{s+ \beta(2^{-ku}(r+1) )^{\alpha} }(\Omega)} \lesssim 2^{k(s-\frac{d}{p} )}. 
 \end{equation}
 Thus, from (\ref{lem5-1-2}), (\ref{lem5-1-3}), and (\ref{lem5-1-4}), we obtain the following: 
 \begin{equation}\label{lem5-1-5}
 \left(\int_{2^{-ku}}^{1}{(\omega_{r, p}^{*}(f, t) )^{q}\bunsu{1}{t}} \mathrm{d}t \right)^{\frac{1}{q}} \lesssim 2^{k(s-\frac{d}{p} )}. 
 \end{equation}
Therefore, from (\ref{lem5-1-1}) and (\ref{lem5-1-5}),  $| f |_{B_{p, q}^{s(x)}(\Omega)}\lesssim 2^{k(s-\frac{d}{p} )}$. Since $\|f\|_{p} \asymp 2^{-k \frac{d}{p}} \leq 2^{k(s-\frac{d}{p} )}$, we obtain the desired result. 
\qed
\end{Proof}

\end{lemma}

By using \autoref{bsp-hyouka}, we prove \autoref{linear-superior}.
\begin{Proof}
We follow the argument of Theorem\ 1 in \cite{zhang2002wavelet}, and Theorems\ 1 and \ 6 in \cite{suzuki2019deep}. We summarize the argument of Theorem\ 1 in \cite{zhang2002wavelet}. \\
 Let the partition that divides $\Omega$ into cubes, the lengths of which are $2^{-k}$, be $\mathcal{B}$. That is, let 
\[ B_{j, k}= \prod_{i=1}^{d} [ 2^{-k} (j_{i} -1), 2^{-k} j_i ] \hspace{1cm} ( j_i \in \sizen, 1\leq j_i \leq 2^{k}, i=1, \ldots, d) \]
 and $\mathcal{B}= \bigcup_{1\leq j_i \leq 2^{k} } B_{j, k}$. We also let $A_{j, k} = \{ i; X_i \in B_{j, k} \}$ and $| A_{j, k}| = \mathrm{card}( A_{j, k} ) $. If for $0 < \alpha < \beta < 1$, $k\in \sizen$ satisfies $n^{\alpha} < 2^{kd} < n^{\beta}$, there exist some constants $C, C^{'} > 0$, and the following two conditions hold:
 \begin{align}
 &1.\ |\{ x_i \mid x_i \in B\ ( i\in \{1, \ldots, n\} ) \}| \leq C \frac{n}{2^{kd}} \quad(^\forall B \in \mathcal{B}), \notag \\
 &2.\ \mathcal{D} = \{ | A_{j, k} | \leq C^{'} \frac{n}{2^{kd} }; 1\leq j_i \leq 2^{k}, i=1, \ldots, d \} , P(\mathcal{D})=1+o(1). \notag \\ \notag
\end{align}
 Additionally,  let $\mathcal{F}^{\circ}$ be the function set on $\Omega$ and satisfy the following conditions for $\Delta >0$:
 \begin{align}
 1.\ &\mbox{There exists}\ F > 0, \mbox{and for all}\ B_{j, k} \in \mathcal{B}, \mbox{there exists } g \in \mathcal{F}^{\circ}\ \mbox{that satisfies } \notag \\
 &g(x) \geq \Delta F\quad(^\forall x\in B_{j, k} ), \notag\\
 2.\ &\mbox{There exists}\ C^{''} >0 \ \mbox{and}\ f\in F^{\circ} \ \mbox{satisfies}\ \frac{1}{n} \sum_{i=1}^{n} f(x_i) ^2 \leq C^{''}\Delta^2 2^{-kd} \notag\\ 
 &\mbox{on the event}\ \mathcal{D}. \notag
 \end{align}
Let the conditions above be condition A. In addition, we let 
 \[ R^{*} = \inf_{\hat{f}: \mathrm{linear}}\sup_{ f\in \mathcal{G} }\expect[ \| f -\hat{f} \|_{L_{2}(P_X)} ^2 ].\]
For $0 < \alpha < \beta < 1$, suppose that $k\in \sizen$ satisfies $n^{\alpha} < 2^{kd} < n^{\beta}$ and there exists the function set $\mathcal{F}^{\circ}$ that satisfies condition A. Then, there exists a constant $ F_1 > 0$ such that at least one of the following inequalities hold for a sufficiently large $n$: 
 \begin{align}
 \frac{F^2}{4F_1 C^{''}}\frac{2^{kd}}{n} &\leq R^{*}, \notag \\
 \frac{F^3}{32} \Delta^2 2^{-kd} &\leq R^{*}. \notag
 \end{align}
 The argument above follows that in \cite{zhang2002wavelet}. By using this, we prove the theorem.
 
 Let $\Delta=2^{-k(s-\frac{d}{p}) }$. In addition, we let $\omega = (\omega_j )_{j\in \Lambda(k)}$ be a one-hot vector. That is, there exists $ j\in \Lambda(k)$ that satisfies $\omega_j =1$ and $\omega_{j^{'}}= 0\ (^\forall j^{'}\neq j)$. We define $f_{\omega} $ as follows:
 \[ f_{\omega} \coloneqq \sum_{ j\in J(k) } \Delta \omega_j M_{k, j}^d (x). \]
 We take a sufficiently large $k$, and for each $\omega_j$, let $c_j \in \mathrm{supp}\ M_{k, j}^d$. By the argument in \autoref{bsp-hyouka}, there exists a constant $C_1$ that does not depend on $j$ and $k$, and it holds that $f_{\omega_j} \subset C_1 U _{B_{p, q}^{s+\beta\| x-c_j \|_2 ^{\alpha} }(\Omega)}$. It is clear that there exists $f_{\omega}$ that satisfies condition A-1. Condition A-2 is satisfied because for any $\omega$, there exists $B \in \mathcal{B}$ such that
 \[ \frac{1}{n} \sum_{i=1}^{n} f_{\omega}(x_i) ^2 \lesssim \frac{1}{n} \Delta^2 | \{ i \mid x_i \in B \ (i=1, \cdots n)  \} | \leq C^{''}\Delta^2 2^{-kd} \]
on the event $\mathcal{D}$. If we take $k$ that satisfies $n\asymp 2^{k(2(s-\nu)+d) }$, it holds that
\[ R^{*} \gtrsim n^{-\frac{2(s-\nu)}{2(s-\nu)+d}}.\]
 
\qed 
\end{Proof}

\end{document}